\documentclass[a4paper,BCOR=.25mm]{scrartcl}

\usepackage[margin=2.5cm
]{geometry}

\usepackage[utf8]{inputenc}
\usepackage[T1]{fontenc}

\usepackage[english]{babel}
\usepackage{amsmath}
\usepackage{amssymb}
\usepackage{mathtools}
\usepackage{booktabs}
\usepackage{caption}
\usepackage{subcaption}
\usepackage{multirow}
\usepackage{xstring}
\usepackage{xcolor}

\DeclareMathOperator{\Hel}{Hel}
\newcommand{\dif}{\mathop{}\!\mathrm{d}}
\newcommand{\nicebold}{\fontseries{b}\selectfont}
\newcommand{\mdet}[1]{\ensuremath{\mathrm{det}(#1)}}
\newcommand{\bs}[1]{\ensuremath{\boldsymbol{#1}}}

\newcommand{\measureformat}[1]{{\fontfamily{cmss}\selectfont\textbf{#1}}}
\newcommand{\measure}[1]{\measureformat{\resolvemeasure{#1}}}
\newcommand{\resolvemeasure}[1]{\IfEqCase{#1}{{Info}{Informativeness}
        {info}{informativeness}
        {Uniq}{Uniqueness}
        {uniq}{uniqueness}        
        {Impo}{Importance}
        {impo}{importance}
        {Disc}{Discrimination}
        {disc}{discrimination}
        {Repr}{Representativity}
        {repr}{representativity}
        {Unct}{Uncertainty}
        {unct}{uncertainty}
        {Dsng}{Distinguishability}
        {dsng}{distinguishability}}[#1]}
\newcommand{\measureabrv}[1]{\measureformat{\resolvemeasureabrv{#1}}}
\newcommand{\resolvemeasureabrv}[1]{\IfEqCase{#1}{
        {Unct}{Unce}
        {unct}{unce}
        {Dsng}{Dsng}
        {dsng}{dsng}}[#1]}

\newcommand*\samethanks[1][\value{footnote}]{\footnotemark[#1]}

\usepackage[babel]{microtype}
\begin{document}

  \title{Towards Automation of Knowledge Understanding: An Approach for Probabilistic Generative Classifiers}
  \date{}

  \author{Dominik Fisch \thanks{BMW Group, 80788 Munich, Germany (e-mail: dominik.fisch@bmw.de)} \and
     Christian Gruhl \thanks{University of Kassel, Department of Electrical Engineering and Computer Science, Wilhelmshoeher Allee 73, 34121 Kassel, Germany (e-mail: \{cgruhl,kalkowski,bsick\}@uni-kassel.de)} \and
     Edgar Kalkowski \samethanks \and
     Bernhard Sick \samethanks \and
     Seppo J. Ovaska \thanks{Aalto University, Department of Electrical Engineering and Automation, Espoo, Finland (e-mail: seppo.ovaska@aalto.fi)}}

  \maketitle


  \begin{abstract}
    After data selection, pre-processing, transformation, and feature extraction, knowledge extraction is not the final step in a data mining process.
    It is then necessary to understand this knowledge in order to apply it efficiently and effectively.
    Up to now, there is a lack of appropriate techniques that support this significant step.
    This is partly due to the fact that the assessment of knowledge is often highly subjective, e.g., regarding aspects such as novelty or usefulness.
    These aspects depend on the specific knowledge and requirements of the data miner.
    There are, however, a number of aspects that are objective and for which it is possible to provide appropriate measures.
    In this article we focus on classification problems and use probabilistic generative classifiers based on mixture density models that are quite common in data mining applications.
    We define objective measures to assess the informativeness, uniqueness, importance, discrimination, representativity, uncertainty, and distinguishability of rules contained in these classifiers numerically.
    These measures not only support a data miner in evaluating results of a data mining process based on such classifiers.
    As we will see in illustrative case studies, they may also be used to improve the data mining process itself or to support the later application of the extracted knowledge.
  \end{abstract}



\section{Introduction} \label{sec_intro}

\begin{figure}
  \centering
  \includegraphics{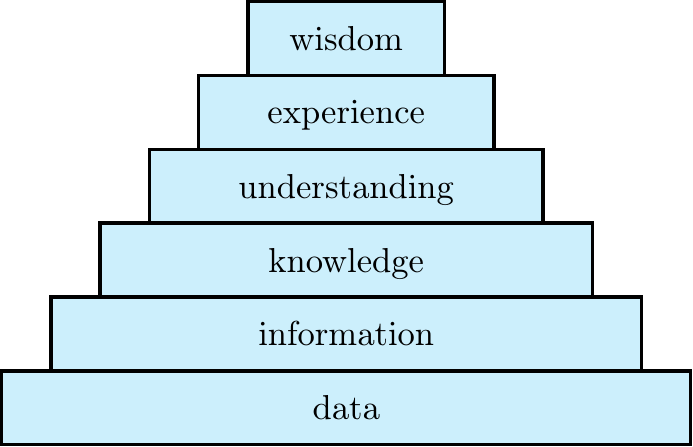}
  \caption{The data mining pyramid (adopted from \cite{Embrechts}).\label{fig-pyramid}}
\end{figure}

Data mining (DM) can be seen as a multi-step process as shown in the data mining pyramid (see Fig.\ \ref{fig-pyramid}) which was introduced by Embrechts et al.\ in \cite{Embrechts}.
The idea of this pyramid can briefly be summarized as follows: Raw \textit{data} are pre-processed to condense application-specific \textit{information} in attributes or features.
Then, \textit{knowledge} is extracted, e.g., by building classification or regression models.
By analyzing this knowledge off-line (i.e.,~after the model is learned from training data) and by using it in a given application (on-line) it is possible to come to a deeper \textit{understanding} of its working principles and to gain some \textit{experience} in using it, respectively.
Both will support the efficient and effective application of the knowledge.
Finally, this kind of meta-knowledge (knowledge about knowledge) will eventually help to solve similar kinds of application problems.
That is, the final step of \textit{wisdom} is reached by transferring the knowledge to other application domains.
While the steps from data over information to knowledge are well supported by appropriate algorithms and commercial or free tools, there is still a lack of techniques that support the subsequent steps.

In this article, we address the problem of knowledge understanding by analyzing its properties off-line.
We use \textit{classifiers based on probabilistic mixture models} (CMM, see \cite{FS09,Bis06}) which can be used in many DM applications.
These classifiers can be termed to be hybrid in the sense that different kinds of distributions are combined for different kinds of attributes (e.g., continuous or categorical) which makes the classifier very flexible.
CMM are trained from data samples using expectation maximization or related techniques such as variational Bayesian approaches \cite{Bis06}.
For CMM, we propose various measures that assess the \textit{informativeness}, \textit{uniqueness}, \textit{importance}, \textit{discrimination}, \textit{representativity}, \textit{uncertainty}, and \textit{distinguishability} of rules contained in this classifier.
The measures can be used in different ways: For example, it is possible to prune classifiers, to rank classification rules, or to detect novel kinds of knowledge (cf.\ anomaly detection) during the classifier's application.
However, the proposed measures should be seen as a first yet important step towards an automation of knowledge understanding. Thus, with the assumptions we make, we do not cover all possible aspects of real World applications so far. For instance, other probability distributions are needed for other attribute types.

This article is a substantially extended version of a conference article (cf.\ \cite{FKS11}). The main contributions compared to the previous article are: All existing measures are revised and we introduce two additional measures (\measure{unct} and \measure{dsng}) to further analyze different aspects of generative classifiers. The evaluation of the measures is also completely new and improved. In four case studies utilizing over 20 artificial and real-world benchmark data sets we illustrate how our proposed measures can help data scientists.

In the remainder of the article we first briefly discuss related work in Section~\ref{sec_related}.
Then, we describe the classifier and introduce the various measures in Section \ref{sec_methods}.
Illustrative case studies in Section \ref{sec:CaseStudies} show how the new measures can be applied.
In Section~\ref{sec_conclusion} we summarize the key findings and give an outlook to future work.

\section{Related Work} \label{sec_related}

Obviously, the measures we are looking for are closely related to so-called \textit{interestingness measures} in data mining.
\textit{Data mining (DM)}, today often used as a synonym of \textit{knowledge discovery in databases (KDD)}, deals with the ``the nontrivial process of identifying valid, novel, potentially useful, and ultimately understandable patterns in data'' \cite{Fayyad1996}.
But, how can this ``interestingness'' be assessed numerically? Obviously, there are objective facets of interestingness such as validity and subjective facets such as usefulness (see, e.g., \cite{Hilderman2001b,Hilderman2001,McGarry2005,Shekar2004,Tuzhilin2002}).

\textit{Objective measures} analyze the extracted knowledge without relating it to the users' prior knowledge or needs.
These measures are based, e.g., on so-called information criteria or on data-based measurement techniques (see  \cite{Vas14} for an overview).
Examples for the former are Akaike's information criterion or the Bayesian information criterion.
Examples for the latter are statistical measures such as sensitivity, specificity, precision, etc.\ determined with a cross-validation or bootstrapping method on test data (see, e.g., \cite{LMVL08,SHD11,TGTB13,KB09,MO13,HZ10}).
Some other criteria assess the complexity of rules or rule sets such as a rule system size measure (depends on the number of rules), a computational complexity measure (the CPU run-time needed to evaluate a rule or a rule system), a rule complexity measure (number of attributes considered by a rule), a mean scoring rules measure (the average number of rules that are evaluated to find a conclusion), a fuzzy quality measure (for linguistic terms associated with rules), or the information gain for association rules {(see, e.g., \cite{ABP04,HC07,JS03,DHP06}). 
Sometimes, several measures are combined \cite{ABP04,TG97}.
In \cite{GSS12}, interestingness measures for rules are evaluated with regard to the four properties of confirmation, locality, symmetry, and a property termed $\text{Ex}_1$ which assures that conclusively confirmatory rules are assigned a higher interestingness value than non-conclusively confirmatory rules and vice versa.
Especially, in \cite{GSS12} weaker forms of the locality property and $\text{EX}_1$ are proposed together with a new interestingness measure that fulfills the weaker forms of those properties.
Two further measures are proposed in \cite{Gla14} and compared to the measure from \cite{GSS12} in terms of the properties they fulfill.
Also, in \cite{Gla14} two new Bayesian confirmation measures for the evaluation of rule interestingness measures are proposed.
An interestingness measure that does not evaluate rules or classifier components but individual samples is presented in \cite{BFS11}.
There, a support vector machine is used and samples close to the decision boundary or distant to any previous samples are classified as interesting.
In \cite{WLHL10}, a support vector machine is used to learn interestingness values of Twitter hashtags from data.

\textit{Subjective measures} consider additional knowledge about the application and information about the data miner such as skills and needs \cite{Shapiro1994,Padmanabhan1999}.
Examples for subjective measures are novelty \cite{Basu2001,Fayyad1996}, usefulness \cite{Fayyad1996}, understandability \cite{Fayyad1996}, actionability \cite{CZZL10,Cao10}, and unexpectedness \cite{Fiore2002,Jaroszewicz2004,Suzuki2002,PT06}.

Existing measures are based on different techniques to represent information about the human users and they depend on the form of knowledge representation.
Often, Bayesian networks, fuzzy classifiers, or association rules are addressed in related work.

Related work can also be found in the field of \textit{recommender systems} (e.g., in a content-based approach or a collaborative filtering approach) \cite{Adomavicius2005a,Herlocker2004}.

In  this article, we focus on \textit{objective} measures beyond existing measures for validity (e.g., precision, recall, etc.).
We also focus on measures for classification rules and rule sets that are not ``crisp'', but consider samples and knowledge that are ``uncertain''.
More specifically, we consider probabilistic classifiers and stay within a probabilistic methodological framework to assess the rules contained in these classifiers.
In Section \ref{sec_conclusion} we will also address the question of whether these measures could be transferred to other kinds of classifiers, too.

\section{Methodological Foundations} \label{sec_methods}

In this section we will first present the generative classifier paradigm.
A \emph{generative} classifier aims at modeling the processes underlying the ``generation'' of the data \cite{Bis06}.
It is termed to be ``generative'' because  if these processes are modeled perfectly, a generative model could be used to generate artificial data with exactly the same characteristics as the real data.
In contrast, discriminative classifiers aim at finding the optimal decision boundary directly.
Today, these two approaches are often combined to exploit their respective advantages.
Here, we use probabilistic techniques for our generative classifiers.
In a second part of the section, we will introduce our new measures for knowledge understanding.

\subsection{Probabilistic Classifier CMM} \label{subsec:cmm}


The classifiers we are using here are probabilistic classifiers.
That is, for a given, specific $D$-dimensional input sample $\mathbf{x}'$ we want to compute the posterior distribution $p(c|\mathbf{x}')$, i.e., the probabilities for class membership (with classes $c \in \cal C$) given the input sample $\mathbf{x}'$.
To minimize the risk of classification errors we may then select the class with the highest posterior probability (cf.\ the principle of \emph{winner-takes-all}).
Generally, the posterior distribution $p(c|\mathbf{x})$ can be determined by (cf.\ \cite{FS09})
\begin{equation}
  \label{eq_cmm_eins}
  \begin{split}
    p(c|\mathbf{x}) =
    \frac{p(\mathbf{x}|c) p(c)}{p(\mathbf{x})} =
    \frac{p(c) \sum_{i \in \cal I} p(\mathbf{x}|c,i) p(i|c)}{p(\mathbf{x})},
  \end{split}
\end{equation}
where $p(c)$ is a multinomial distribution with parameters that are termed {\em class priors} and conditional densities $p(\mathbf{x}|c,i)$ that are called {\em components}.
$\cal I$ is the overall set of components in this model.
In an approach with separate sets of components for the different classes (i.e., we uniquely assign the components to classes, $p(c|i) \in \{0,1\}$), we can conclude that we only have to sum up over all components assigned to a certain class to determine the class posteriors:
\begin{equation}
  \label{eq_cmm_zwei}
  \begin{split}
    p(c|\mathbf{x}) = \frac{\sum_{i \in {\cal I}_c} p(\mathbf{x}|i_c) p(i_c)}{p(\mathbf{x})},
  \end{split}
\end{equation}
where ${\cal I}_c$ (with ${\cal I}_c \subset \cal I$) is the set of components assigned to class $c \in \cal C$.
Note that
\begin{equation}
  p(\mathbf{x}) = \sum_{c \in \cal C} \sum_{i_c \in {\cal I}_c}  p(\mathbf{x}|i_c) p(i_c) = \sum_{i \in \cal I}  p(\mathbf{x}|i) p(i).
  \label{eqn:px}
\end{equation}
The parameters $\pi_i$ of the multinomial distribution $p(i)$ are called {\em mixing coefficients}.

Altogether, we have a classifier consisting of a linear combination of components, where each component is described by a distribution $p(\mathbf{x}|c, i)$.
To keep the notation uncluttered, a specific component is identified by a single index $i \in \cal I$ in the following (i.e., $p(\mathbf{x}|i)$) if its class is not relevant.
Which kind of density functions can we use for the components? 
Considering a $D$-dimensional sample $\mathbf{x}$ it may have $D_\textrm{cont}$ continuous
(i.e., real-valued) dimensions (attributes, features) and $D_\textrm{cat}=D - D_\textrm{cont}$ categorical ones.
Without loss of generality we arrange these dimensions such that
\begin{equation}
  \mathbf{x} = (\underbrace{x_1,\dots,x_{D_\textrm{cont}}}_{\textrm{continuous}},\underbrace{\boldsymbol{x}_{D_{\textrm{cont}}+1},\dots,\boldsymbol{x}_D}_{\textrm{categorical}}).
\end{equation}
Note that we italicize $x$ when we refer to single dimensions.
The continuous part of this vector $\mathbf{x}^\textrm{cont} = (x_1,\dots,x_{D_\textrm{cont}})$ with $x_d \in \mathbb{R}~$ for all $d \in \{1,\dots,D_\textrm{cont}\}$ is modeled with a multivariate \emph{normal} (i.e., Gaussian) distribution with center $\boldsymbol{\mu}$ and covariance matrix $\boldsymbol{\Sigma}$.
That is, with $\mathrm{det}(\cdot)$ denoting the determinant of a matrix we use the model
\begin{equation}
  \label{eqn:gauss}
  \begin{split}
    {\cal N}(\mathbf{x}^\textrm{cont}|\boldsymbol{\mu},\boldsymbol{\Sigma}) =
    \frac{1}{(2\pi)^{\frac{D_\textrm{cont}}{2}} \mdet{\boldsymbol{\Sigma}}^{\frac{1}{2}}} \exp\left(-\frac{1}{2} (\mathbf{x}^\textrm{cont} - \boldsymbol{\mu})^{\mathrm{T}} \boldsymbol{\Sigma}^{-1} (\mathbf{x}^\textrm{cont} - \boldsymbol{\mu})\right)
  \end{split}
\end{equation}
For many practical applications, the use of Gaussian components or Gaussian mixture models (GMM) can be motivated by the generalized \textit{central limit theorem} (cf., e.g., \cite{DHS01}) which roughly states that the sum of independent samples from any distribution with finite mean and variance converges to a normal distribution as the sample size goes to infinity. Moreover, any continuous distribution can be approximated arbitrarily well by a finite mixture of normal densities \cite{McLP00}.

For categorical dimensions we use a 1-of-$K_d$ coding scheme where $K_d$ is the number of possible categories of attribute $\boldsymbol{x}_d$ ($d \in \{D_{\textrm{cont}}+1,\dots,D\}$).
The value of such an attribute is represented by a vector $\boldsymbol{x}_d = (x_{d_1},\dots,x_{d_{K_d}})$ with $x_{d_k} = 1$ if $\boldsymbol{x}_d$ belongs to category $k$ and $x_{d_k} = 0$ otherwise.
The classifier models categorical dimensions by means of a special case of \emph{multinomial} distributions.
That is, for an input dimension (attribute) $\boldsymbol{x}_d \in \{\boldsymbol{x}_{D_{\textrm{cont}}+1},\dots,\boldsymbol{x}_D\}$ we use
\begin{equation}
  {\cal M}(\boldsymbol{x}_d|\boldsymbol{\delta}_d) = \prod_{k=1}^{K_d} \left(\delta_{d_k}\right)^{x_{d_k}}
\end{equation}
with parameters $\boldsymbol{\delta}_d = (\delta_{d_1},\dots,\delta_{d_{K_d}})$ and restrictions $\delta_{d_k} \geq 0$ and $\sum_{k=1}^{K_d}\delta_{d_k} = 1$.

We assume that the categorical dimensions are mutually independent and that there are no dependencies between the categorical and the continuous dimensions.
Then, the component densities $p(\mathbf{x}|i)$ are defined by
\begin{equation}
  \label{eqn:hybridcomponent}
  p(\mathbf{x}|i) = {\cal N}(\mathbf{x}^\textrm{cont}|\boldsymbol{\mu}_{i}, \boldsymbol{\Sigma}_{i}) \cdot
  \prod_{\mathclap{d=D_\textrm{cont}+1}}^D
  {\cal M}(\boldsymbol{x}_d|\boldsymbol{\delta}_{d_i}).
\end{equation}
With this approach it is possible to model multivariate categorical data sets arbitrarily well despite any independence assumption concerning the categorical variables.

Other dimensions (i.e.,~other feature types that are not continuous or categorical) can be handled by relying on other (hybrid) distributions, which are not considered yet. Beside that, for some feature types conversions under certain assumptions are possible. For instance, rational ($\mathbb{Q}$) and integer ($\mathbb{N}$) dimensions might be interpreted as nearly continuous under appropriate conditions (e.g.,~if their support set is large enough). Other attributes (e.g.,~integers with few values actually occurring) can be categorized as preprocessing step before the CMM is trained.

In Section \ref{sec:classifierscomp} we compare CMM to some other classifier paradigms regarding classification accuracy. In general, these probabilistic generative classifiers offer some other interesting features: risk minimizing cost functions can easily be combined with probabilistic outputs, class priors can be compensated, different models can easily be combined, or anomaly detection techniques can be defined \cite{Bis06,TAAS2011}.

\subsubsection{Training of CMM} \label{sec:TrainingOfCMM}

How can the various parameters of the classifier be determined? For a given training set $\mathbf{X}$ with $N$ samples $\mathbf{x}_n$ it is assumed that the $\mathbf{x}_n$ are independent and identically distributed (i.i.d.).
First, $\mathbf{X}$ is split into $C$ subsets $\mathbf{X}_c$, each containing all samples of the corresponding class $c$, i.e.,
\begin{equation}
 \mathbf{X}_c = \{\mathbf{x}_n|\mathbf{x}_n \text{ belongs to class } c\}.
\end{equation}
For each $\mathbf{X}_c$, a mixture model is trained.
Here, we perform the parameter estimation by means of a technique called \textit{variational Bayesian inference} (VI) which realizes the Bayesian idea of regarding the model parameters as random variables whose distributions must be trained \cite{FS09}.
This approach has two important advantages over other methods such as a standard expectation maximization (EM) approach.
First, the estimation process is more robust, i.e., it avoids ``collapsing'' components, so-called singularities when the variance in one or more directions vanishes.
Second, VI optimizes the number of components on its own by pruning irrelevant components, i.e., those that are not considered to notably contribute to the overall density (which is reflected by a reasonably small mixing coefficient). Therefore we start the training with a relatively large number of components and rely on VI to  automatically reduce the number of components to a sufficient number. That is, the number of model components is not a critical parameter that has to be considered by a user.
For a more detailed discussion on Bayesian inference, and, particularly, VI see \cite{Bis06}.
More details concerning the training algorithm can be found in \cite{FS09}.

At this point, we have obtained parameter estimates for the $p(\mathbf{x}|c,i)$ and $p(i|c)$, cf.\ Eq.\ \eqref{eq_cmm_eins}.
The parameters for the class priors $p(c)$ are estimated with
\begin{equation}
  \gamma_c = \frac{|\mathbf{X}_c|}{|\mathbf{X}|}
\end{equation}
where $|\cdot|$ denotes the cardinality of a set.

It should be noted that most of the measures defined in this article are independent from the specific training technique, e.g., VI or EM.

\subsubsection{Classification Performance of CMM Compared to 1NN, SVM, and Decision Trees}\label{sec:classifierscomp}

To confirm that CMM are comparable to other classification paradigms regarding their classification accuracy and, thus, sufficiently meaningful to define measures for knowledge understanding, we evaluate their performance on 21 benchmark data sets:
phoneme, satimage (real-world data from the UCL Machine Learning Group \cite{UCL14});
australian, credit\_a, credit\_g, ecoli, glass, heart, iris, pendigits, pima, quality, seeds, segment, vehicle, vowel, wine, yeast (real-world data from the UCI Machine Learning Repository \cite{AN07});
clouds (artificial data from the UCI Machine Learning Repository \cite{AN07});
ripley (artificial data proposed in \cite{Rip96});
and two\_moons (own, artificial data set).
Table~\ref{tab:DataSetsBasic} contains general information about all data sets, e.g., number of samples, feature (attribute) types, and class distribution.

\begin{table}[htb]
  \centering\scriptsize
  \caption{General information about 21 benchmark data sets.}
  \label{tab:DataSetsBasic}
  \begin{tabular}{l c c c c c}
    \toprule
    \multirow{2}{*}{Data Set} & Number\ of & Continuous  & Categorical & Number\ of & Class                                            \\
                              & Samples    & Attributes\ & Attributes  & Classes    & Distribution (in \%)                             \\
    \midrule
    australian                & 690        & 6           & 8           & 2          & 55.5,44.5                                        \\
    clouds                    & 5000       & 2           & --          & 2          & 52.2,50.0                                        \\
    credit\_a                  & 690        & 6           & 9           & 2          & 44.5,55.5                                        \\
    credit\_g                 & 1000       & 7           & 13          & 2          & 70.0,30.0                                        \\
    ecoli                     & 336        & 7           & --          & 8          & 42.6,22.9,15.5,10.4,5.9,1.5,0.6,0.6              \\
    glass                     & 214        & 9           & --          & 6          & 32.7,35.5,7.9,6.1,4.2,13.6                       \\
    heart                     & 270        & 6           & 7           & 2          & 44.4,55.6                                        \\
    iris                      & 150        & 4           & --          & 3          & 33.3,33.3,33.3                                   \\
    pendigits                 & 10992      & 16          & --          & 10         & 10.4,10.4,10.4,9.6,10.4,9.6,9.6,9.6,10.4,9.6,9.6 \\
    phoneme                   & 5404       & 5           & --          & 2          & 70.7,29.3                                        \\
    pima                      & 768        & --          & 8           & 2          & 65.0,35.0                                        \\
    quality                   & 4898       & 13          & --          & 7          & 0.4,3.3,29.7,44.9,17.9,3.7,0.1                   \\
    ripley                    & 1250       & 2           & --          & 2          & 50.0,50.0                                        \\
    satimage                  & 6345       & 5           & --          & 6          & 24.1,11.1,20.3,9.7,11.1,23.7                     \\
    seeds                     & 210        & 7           & --          & 3          & 33.3,33.3,33.3                                   \\
    segment                   & 2310       & 19          & --          & 7          & 14.3,14.3,14.3,14.3,14.3,14.3,14.3               \\
    two\_moons                & 14977      & 2           & --          & 2          & 49.1,50.9                                        \\
    vehicle                   & 846        & 18          & --          & 4          & 23.5,25.7,25.8,25.0                              \\
    vowel                     & 528        & 10          & --          & 11         & 9.1,9.1,9.1,9.1,9.1,9.1,9.1,9.1,9.1,9.1,9.1      \\
    wine                      & 178        & 13          & --          & 3          & 33.1,39.8,26.9                                   \\
    yeast                     & 1484       & 8           & --          & 10         & 16.4,28.1,31.2,2.9,2.3,3.4,10.1,2.0,1.3,0.3      \\
    \bottomrule
  \end{tabular}
\end{table}

With the training algorithm Sequential Minimal Optimization (SMO)  \cite{Platt98}, Support Vector Machines (SVM) are one of the most widely used discriminative approaches for pattern recognition tasks. Consequently, we train SVM with the frequently used \textsf{libsvm} library using RBF (or Gaussian) kernels \cite{libsvm}.
Another common paradigm are decision trees (DT). Familiar training algorithms are ID3 or its successor C4.5 \cite{Quinlan1993}, for instance. Here we use the latter to train DT. Among the advantages of a DT trained with C4.5 is that the tree structure can easily be used to extract human-readable rules (e.g.,~\cite{Kretschmann01}).
The last competitor is the simple one-nearest-neighbor classifier (1NN). Even though it only classifies with respect to the nearest neighbor, it can be shown that for $N\rightarrow\infty$ the maximum classification error is at most twice the maximum of a classifier that yields the best possible classification \cite{CoverHart67}.

\begin{table}[htb]
	\caption{Classification accuracy (in \%) for four classifier paradigms on 21 data sets.\label{tab:evaluationValidation}}
	\scriptsize\centering
	\begin{tabular}{l c c c c c }
		\toprule
		Data Set & $\text{CMM}$ & $\text{SVM}$ & DT & 1NN \\
		\midrule
		australian & \nicebold{84.1} & 79.7 & 81.9 & 79.7\\
		clouds & \nicebold{89.5} & 74.9 & 88.4 & 84.8\\
		credit\_a & \nicebold{84.8} & 83.3 & 83.3 & 78.3\\
		credit\_g & 68.5 &  \nicebold{75.0} & 69.5 & 63.5\\
		ecoli & 87.9 & \nicebold{90.9} & 84.8 & 81.8\\
		glass & 72.1 & 69.8 & 67.4 & \nicebold{83.7}\\
		heart & 81.5 & \nicebold{85.2} & 75.9 & 70.4\\
		iris & \nicebold{96.7} & 93.3 & 93.3 & 93.3\\
		pendigits & \nicebold{99.5} & 99.4 & 96.2 & 99.2\\
		phoneme & 86.0 & 78.3 & 84.1 & \nicebold{89.1}\\
		pima & 68.2 & \nicebold{73.4} & \nicebold{73.4} & 66.9\\
		quality & 57.1 & \nicebold{58.1} & 55.3 & 50.5\\
		ripley & \nicebold{92.0} & 90.0 & 90.0 & 88.0\\
		satimage & \nicebold{88.1} & 87.8 & 87.4 & 84.7\\
		seeds & \nicebold{100.0} & \nicebold{100.0} & 95.2 & 90.5\\
		two\_moons & \nicebold{100.0} & 99.3 & 99.8 & \nicebold{100.0}\\
		vowel & 98.0 & 82.3 & 74.7 & \nicebold{98.5}\\
		wine & 97.2 & 97.2 & 97.2 & 97.2\\
		yeast & \nicebold{54.7} & 51.7 & 51.7 & 50.0\\
		\bottomrule
	\end{tabular}
\end{table}

The performance evaluation for all classifiers is done in the same way. At first, 20\% of the samples are taken from each data set to function as an independent test set. On the remaining 80\%, a parameter search is done using grid search combined with a 4-fold cross validation.
The parameters that perform best over all folds are then selected to determine the performance on the test set.
The results on the independent test sets, achieved with the same parameters as for the training sets, are set out in Table \ref{tab:evaluationValidation}. 
We can see that CMM perform comparably well.

\begin{figure}[htb]
	\centering
	\subcaptionbox{Model with 6 components.\label{fig:CMMDecisionBoundaries:6}}{\includegraphics[width=.3\textwidth]{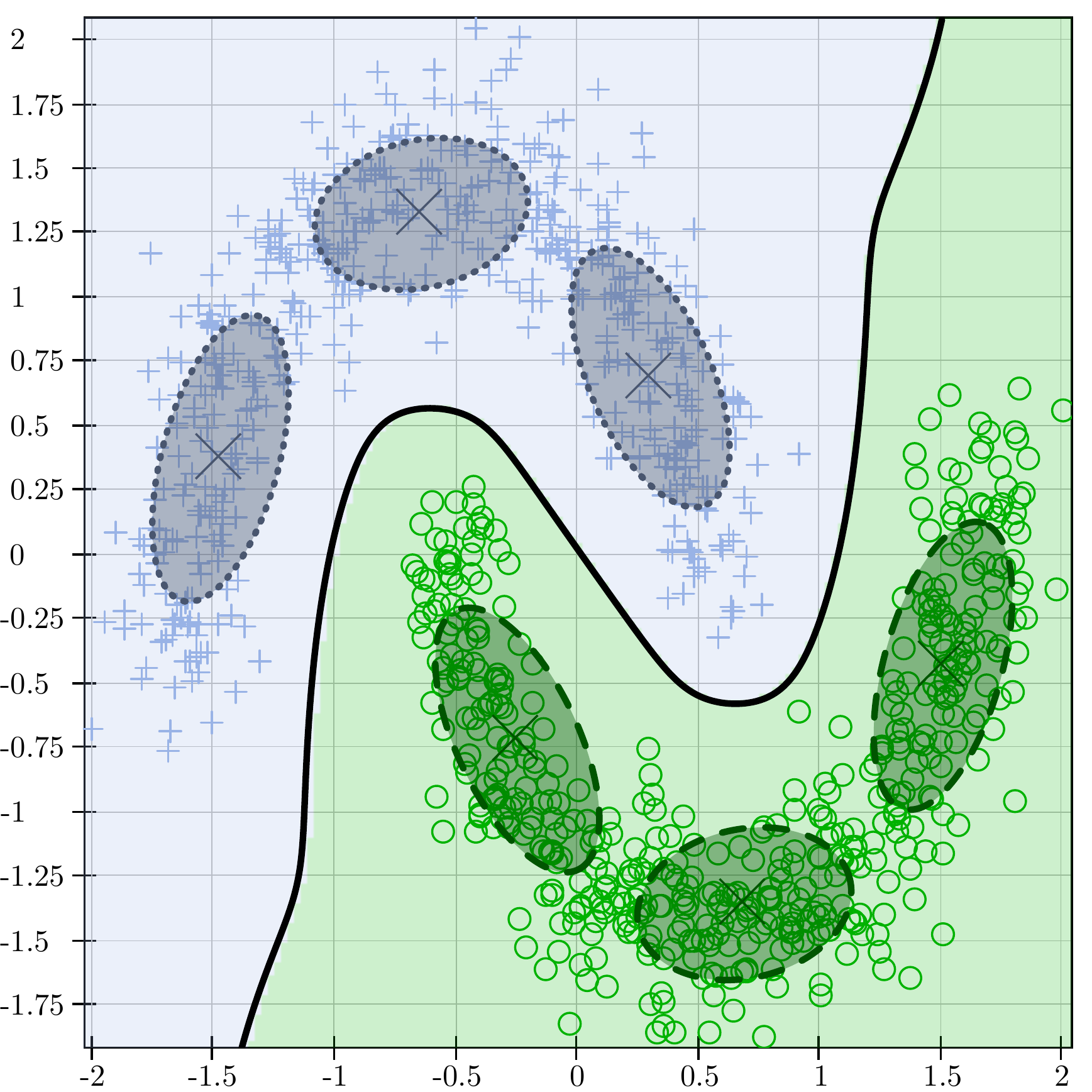}}	
	\subcaptionbox{Model with 8 components.\label{fig:CMMDecisionBoundaries:8}}{\includegraphics[width=.3\textwidth]{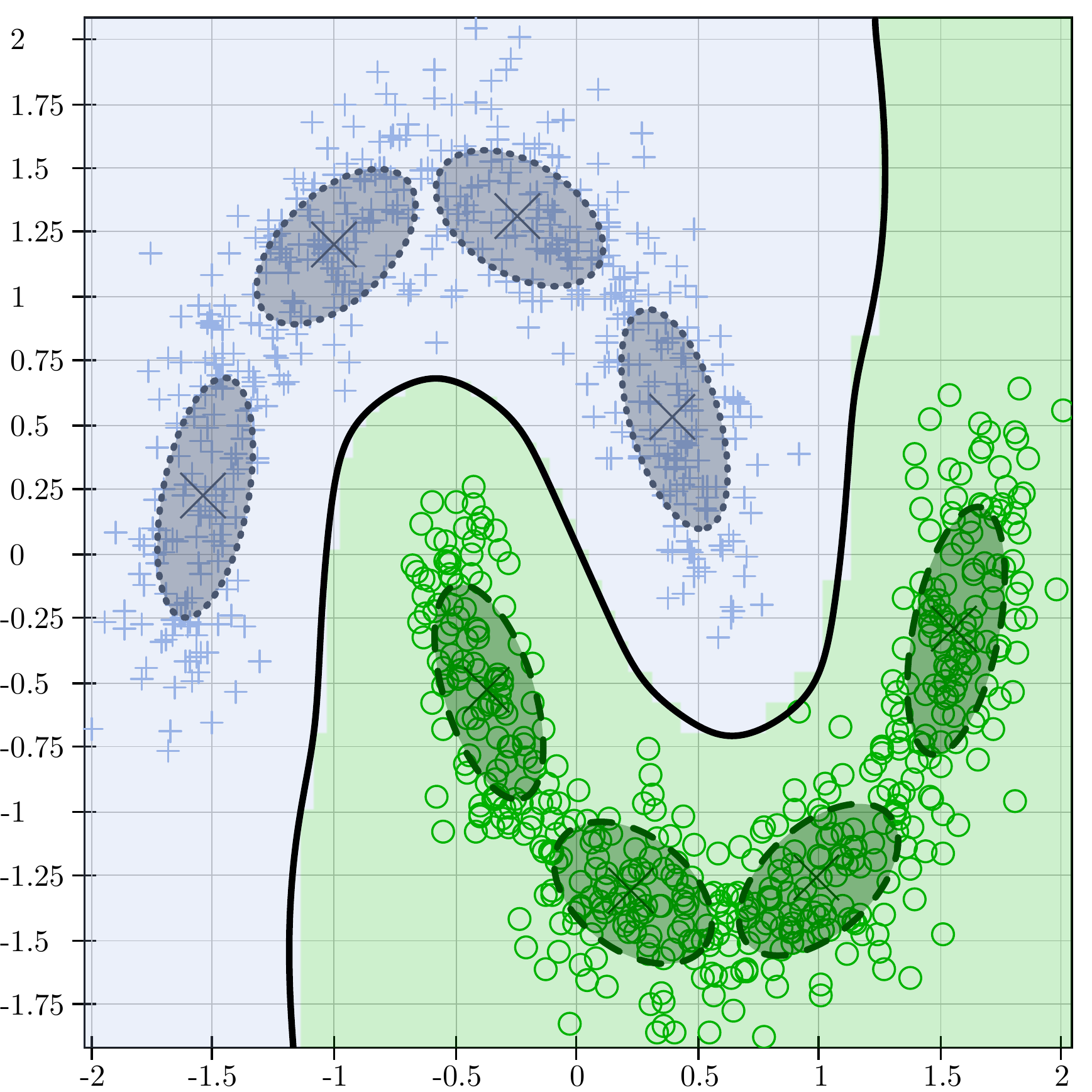}}
	\subcaptionbox{Model with 12 components.\label{fig:CMMDecisionBoundaries:12}}{\includegraphics[width=.3\textwidth]{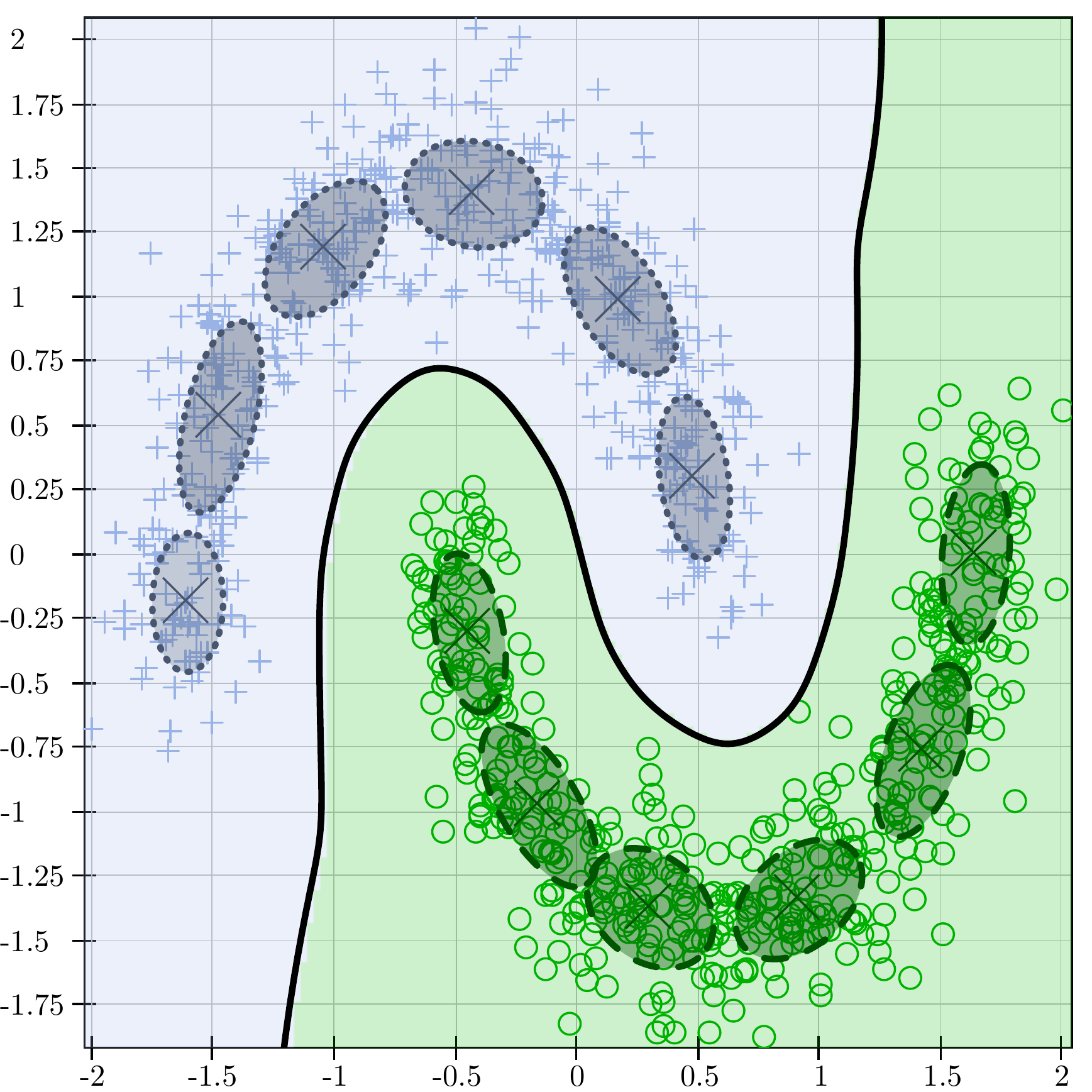}}
	\caption{Different models and classifiers for the artificial \textit{two\_moons} data set, resulting in almost identical decision boundaries. The images show the two-dimensional input spaces of a classifier. Crosses (blue) and circles (green) are samples that originate from processes that can be assigned to two different classes. The probabilistic models being part of the classifiers are depicted by their components (Gaussians) which are represented by ellipses of different shapes (given by the covariance matrices) at positions marked with big $\times$s (given by the expectations). The thick black lines are decision boundaries which separate the regions assigned to different classes.}
	\label{fig:CMMDecisionBoundaries}
\end{figure}

One substantial property of CMM is that the actual number of components in the mixture model has not a strong influence on the decision boundary as long as there is a sufficient number of components in the model. 
Fig.\ \ref{fig:CMMDecisionBoundaries} demonstrates this exemplarily on one of the used data sets (two\_moons). Here, a CMM is trained with VI in such a way that the components of the trained models ``cover'' smaller areas of the input space which leads to models with more components\,---\,6 components in Fig. \ref{fig:CMMDecisionBoundaries} (\subref{fig:CMMDecisionBoundaries:6}) up to 12 components in Fig. \ref{fig:CMMDecisionBoundaries} (\subref{fig:CMMDecisionBoundaries:12}). The images show that the decision boundaries of the resulting CMM classifiers are quite similar. 

We exploit this property by initiating our VI training with a higher number of initial components and relying on its pruning capabilities.

\subsubsection{Rule Extraction From CMM} \label{sec:RuleExtraction}

In some applications it is desirable to extract human-readable rules from a trained classifier.
This is possible with our classifier if it is parametrized accordingly.
For the moment, we focus on a single component $p(\mathbf{x}|i)$ and omit the identifying index $i$.
Restrictions concerning the covariance matrix $\mathbf{\Sigma}$ or the number of categories are not necessary.
However, to extract a rule set from the CMM, the premises for the continuous input dimensions are obtained from their univariate projections on the axes. This implies that the information about dependencies between different dimensions (i.e., given by the covariances) are lost. Thus, we recommend to force the covariance matrices $\mathbf{\Sigma}$ to be diagonal.
Then, the multivariate Gaussians ${\cal N}(\mathbf{x}^\textrm{cont}|\boldsymbol{\mu}, \mathbf{\Sigma})$ can be split into a product consisting of $D_{\textrm{cont}}$ univariate Gaussians $\psi_{d}$ with $d=1,\dotsc,D_{\textrm{cont}}$. In this case, the univariate Gaussians $\psi_{d}$ are identical to the projections of the corresponding input dimensions.
The categorical dimensions can be simplified by considering only categories whose probability is above a certain case-dependent threshold.

A rule set (where each component is represented by exactly one rule) can then be derived from the classifier as follows: the variables are the input variables $x_d$ (components of the $D$-dimensional input variable $\mathbf{x}$) and the output variable $c$ which represents the class label.
The rule premises are realized by conjunctions of the univariate Gaussians $\psi_{d}$ (i.e.,~their density $\psi_{d}(x_d)$) and conjunctions of settings for the categorical dimensions. The settings themselves are written as disjunctions of those categories that have probabilities above a specified threshold.
The conclusions (i.e., the class memberships) are given by one class, which is derived from the class affiliation of the component in the premise.
The rules have a form which is very similar to that of fuzzy rules, but they have a rather different (i.e., probabilistic) interpretation.

\begin{figure}
  \centering
  \includegraphics{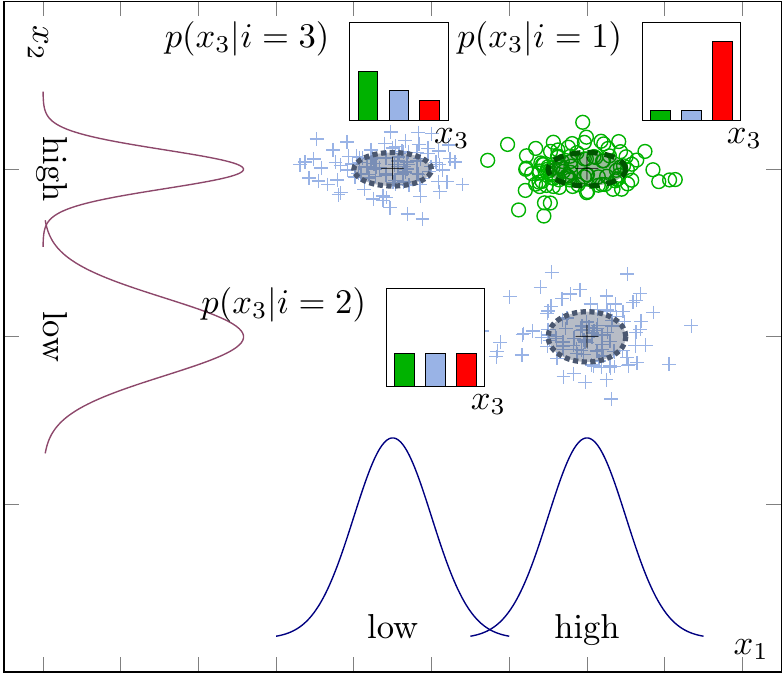}
  \caption{Example of a classifier consisting of three components with two continuous ($x_1,x_2$) and a single categorical dimension ($x_3$). Simplified projections of the continuous densities are depicted at the corresponding axes. The distribution of the categorical dimension is shown as histogram attached to the affiliated component.}
  \label{fig:rules}
\end{figure}

Fig.~\ref{fig:rules} gives an example for a CMM that is used to extract a rule set.
The classifier is embedded in a three dimensional input space with two classes: blue crosses ($+$) and green circles ($\circ$). 
The first two dimensions $x_1$ and $x_2$ are continuous and modeled by three bivariate Gaussians. Their mean values ($\boldsymbol{\mu}_i$) are described by large crosses ($\bs{+}$) and the surrounding ellipses are level curves (i.e.,~surfaces of constant density) of the Gaussians with shapes defined by their covariance matrix $\mathbf{\Sigma}_i$.
Since the covariance matrices $\mathbf{\Sigma}_i$ are only non-zero on their diagonal, all ellipses are axes-oriented.
The corresponding projections onto the axes are also illustrated (i.e.,~the univariate Gaussians). Note, that the projections marked as \textit{high} result from two Gaussians, since the top right component (in green) is covered by the other components (blue) in both continuous dimensions.
The third dimension $x_3$ is categorical with categories A (green), B (blue), and C (red).
The distributions of categories are illustrated by the histograms next to every component.
Here, only categories with a probability strictly greater than the average $1/K_d$ are considered in order to simplify the resulting rules.
Altogether, the following rule set can be extracted from the classifier in Fig.~\ref{fig:rules}:
\begin{quote}
  \begin{minipage}{\textwidth}
    if $x_1$ is \textit{low} and $x_2$ is \textit{high} and ($x_3$ is A or $x_3$ is B)\\
    \hspace*{0.35cm} then c = \textit{blue}\\
    if $x_1$ is \textit{high} and $x_2$ is \textit{high} and $x_3$ is C\\
    \hspace*{0.35cm} then c = \textit{green}\\
    if $x_1$ is \textit{high} and $x_2$ is \textit{low}\\
    \hspace*{0.35cm} then c = \textit{blue}
  \end{minipage}
\end{quote}

Of course, this readability is accomplished at the cost of a limited modeling capability of the classifier (i.e., diagonal covariance matrices and simplified categorical dimensions) and should, thus, only be used if the application requires this kind of human-readable rules.

\subsection{Measures for Knowledge Understanding} \label{sec:Measures}

In the following, we describe seven \textit{objective} measures that can be used to assess the knowledge incorporated in CMM in an objective way.
We will use the term rule instead of component only if we wish to explicitly extract human-readable rules from the CMM.
In this article, we focus on measures for single components (i.e., rules).
Measures for overall classifiers could easily be obtained by averaging measures for components or by considering worst cases etc.
Then, it would also be possible to compare different classifiers, for instance.
In addition, classical performance measures (e.g., classification error on independent test data) should also be used.
If the class a component belongs to is not relevant for a measure, the component is identified by a single index $i \in \{1, \dots, I\}$ with $I = |\cal I|$, e.g., $p(\mathbf{x}|i)$.
Otherwise, it is explicitly denoted by $p(\mathbf{x}|c,i)$.
If sample data are needed to evaluate a measure, we use the training data for that purpose.

\subsubsection{Informativeness} \label{sec:Informativeness}

A component of the CMM is regarded as being very informative if it is assumed to describe a really distinct kind of process ``generating'' data.
To assess the \measure{informativeness} of a component numerically, we use the \textit{Hellinger} distance $\Hel(p(\mathbf{x}),q(\mathbf{x}))$ of two probability densities $p(\mathbf{x})$ and $q(\mathbf{x})$ (cf. \cite{Bis06}).
Compared to other statistical distance measures such as the Kullback-Leibler divergence (cf. Section \ref{sec:Representativity}), the Hellinger distance has the advantage of being bounded between $0$ and $1$.
It is defined by
\begin{equation}
  \Hel(p(\mathbf{x}),q(\mathbf{x})) = \sqrt{1 - \mathrm{BC}(p(\mathbf{x}),q(\mathbf{x}))},
  \label{eq:hellinger}
\end{equation}
where $\mathrm{BC}(p(\mathbf{x}),q(\mathbf{x}))$ denotes the \textit{Bhattacharyya coefficient}:
\begin{equation}
    \mathrm{BC}(p(\mathbf{x}),q(\mathbf{x})) = \int \sqrt{p(\mathbf{x})\cdot q(\mathbf{x})} \dif{x}
\end{equation}

or for discrete distributions with definition range $\mathcal{X}$:
\begin{equation}
\mathrm{BC}(p(\mathbf{x}),q(\mathbf{x})) = \sum_{x \in \mathcal{X}} \sqrt{p(\mathbf{x})\cdot q(\mathbf{x})}
\end{equation}

$\Hel(p(\mathbf{x}),q(\mathbf{x}))$ is $0$ if $p(\mathbf{x})$ and $q(\mathbf{x})$ are identical and it approaches $1$ when $p(\mathbf{x})$ places most of its probability mass in regions where $q(\mathbf{x})$ assigns a probability of nearly zero and vice versa.

Using Fubini's theorem (cf.\ \cite{fubini07,EncMathFubini}) and considering the discrete nature of the multinomial distribution, the Bhattacharyya coefficient of two components $i$ and $i'$ as defined in Eq.~\eqref{eqn:hybridcomponent} can be computed by 
\begin{equation}
  \label{eqn:bcanalytic}
  \begin{split}
    \mathrm{BC} (p(\mathbf{x}|i),p(\mathbf{x}|i')) = {} &
    \int\sqrt{{\cal N}(\mathbf{x}^{\textrm{cont}}|\bs{\mu}_i, \bs{\Sigma}_i) {\cal N}(\mathbf{x}^{\textrm{cont}}|\boldsymbol{\mu}_{i'}, \bs{\Sigma}_{i'})} \dif\mathbf{x}^{\textrm{cont}}\\
    & \cdot \prod_{{d=D_\textrm{cont}+1}}^{D} \sum_{k=0}^{K_d} \sqrt{{\cal M}(\bs{e}_k|\bs{\delta}_{d_i}) {\cal M}(\bs{e}_k|\bs{\delta}_{d_{i'}})}
  \end{split}
\end{equation}
with $\boldsymbol{e}_k$ being the $k$-th row of the $K_d \times K_d$ identity matrix (i.e., we are iterating over all $K_d$ possible categories of dimension $d$).
The integral can be solved analytically for Gaussians yielding
\begin{equation}
  \begin{split}
    & \int\sqrt{{\cal N}(\mathbf{x}^{\textrm{cont}}|\bs{\mu}_i,\bs{\Sigma}_i) {\cal N}(\mathbf{x}^{\textrm{cont}}|\bs{\mu}_{i'},\bs{\Sigma}_{i'})} \dif\mathbf{x}^{\textrm{cont}} \\
    {}={} & \exp\left(-\frac{1}{8} (\bs{\mu}_i - \bs{\mu}_{i'})^{\mathrm{T}} \left(\frac{\bs{\Sigma}_i + \bs{\Sigma}_{i'}}{2}\right)^{-1}(\boldsymbol{\mu}_i - \bs{\mu}_{i'})\right)
    \cdot {\frac{\sqrt[4]{\mdet{\boldsymbol{\Sigma}_i} \cdot \mdet{\bs{\Sigma}_{i'}}}}{\sqrt{\mdet{\frac{\bs{\Sigma}_i + \bs{\Sigma}_{i'}}{2}}}}}.
  \end{split}
\end{equation}

The informativeness of a component $i$ is then determined by its Hellinger distance calculated with respect to the ``closest'' component $i'$ $(i' \neq i)$ contained in the CMM:
\begin{equation}
  \mathrm{\measureabrv{info}}(i) := \min_{i' \neq i}\bigg(\Hel\Big(p(\mathbf{x}|i'), p(\mathbf{x}|i)\Big)\bigg).
\end{equation}
To assess the informativeness of the overall classifier a weighted average of the informativeness values of all components may be used.
The weights can be determined depending on the respective mixing coefficients and the class priors.

The run-time complexity required to evaluate the informativeness of one component is 
\begin{align}
  \mathcal{O}\left(I \cdot \left(D_{\mathrm{cont}}^3 + D_{\mathrm{cat}} \cdot \max_{d \in \{1, \dots, D_{\mathrm{cont}}\}}\{K_d\}\right)\right)
\end{align}
since for each other component in the classifier we have to compute the determinant of its covariance matrix and iterate over all categories of its categorical dimensions.

\subsubsection{Uniqueness} \label{sec:Uniqueness}

The knowledge contained in the components of a CMM should be unambiguous.
This is measured by the \measure{uniqueness} of a component $i$ which reflects to which degree samples belonging to different classes are covered by that component.
Let $\rho_i(\mathbf{x}_n)$ denote the \textit{responsibility} of component $i$ for the generation of sample $\mathbf{x}_n$, i.e.,
\begin{equation}
  \rho_i(\mathbf{x}_n) := \frac{p(\mathbf{x}_n|i) p(i)}{p(\mathbf{x}_n)}. \label{eq:Responsibilities}
\end{equation}
Then, we define the uniqueness of component $i$ by
\begin{equation}
  \mathrm{\measureabrv{uniq}}(i) := \frac{\sum\limits_{\mathbf{x}_n \in \mathbf{X}_c}\rho_i(\mathbf{x}_n)}{\sum\limits_{\mathbf{x}_n \in \mathbf{X}}\rho_i(\mathbf{x}_n)}.
\end{equation}
To evaluate the uniqueness of a whole classifier we may, e.g., compute the weighted average of the individual components’ uniqueness values (e.g., using the mixing coefficients as weights).

The run-time complexity required to evaluate the uniqueness of a component is $\mathcal{O}(N \cdot I \cdot (D_{\mathrm{cont}}^3 + D_{\mathrm{cat}}))$ since we have to evaluate the density of each sample for the whole CMM to get the responsibilities which involves a matrix multiplication and the iteration over all categorical dimensions.

\subsubsection{Importance} \label{sec:Importance}

The \measure{importance} of a component measures the relative weight of a component within the classifier. In general, either a small or a large number of components may be regarded as ``important'', depending on a concrete application. 
Here, a component $i$ is regarded as very important if its mixing coefficient $\pi_i$ is far above the average mixing coefficient $\overline \pi = \frac{1}{I}$.
To scale the importance of a component to the interval $[0,1]$ we additionally use a boundary function that is comprised of two linear functions.
One projects all mixing coefficients that are smaller than the average to the interval $[0,0.5]$ and the other one maps all mixing coefficients that are larger than the average to $[0.5,1]$.
The importance of component $i$ is then computed by
\begin{equation}
  \mathrm{\measureabrv{impo}}(i) :=
  \begin{cases}
    \frac{\pi_i}{2\overline \pi}, & \pi_i \leq \overline \pi \\
    \frac{1-\pi_i}{2(\overline \pi - 1)} + 1, & \pi_i > \overline \pi \\
  \end{cases}.
\end{equation}
Again, to evaluate the importance with regard to a whole classifier we may, e.g., use a weighted average of the importance values of the contained components.

The run-time complexity required to evaluate the importance of one component is $\mathcal{O}(1)$ since it just involves constant time computations with the mixing coefficient of the component.

\subsubsection{Discrimination} \label{sec:Discrimination}

The \measure{discrimination} measure evaluates the influence of a component $i$ on the decision boundary---and, thus, on the classification performance---of the overall classifier.
To calculate the discrimination of component $i$ we create a second CMM by removing $i$ from the original CMM and re-normalizing the mixing coefficients of the remaining components.
Then, we compare the achieved classification error on training data (or on test data where available) of the original CMM (${\cal E}_{\text{with}}$) to the classification error of the CMM without component~$i$ (${\cal E}_\text{without}$):
\begin{equation}
  \mathrm{\measureabrv{disc}}(i) := {\cal E}_\text{without} - {\cal E}_\text{with}.
\end{equation}
If required by the concrete application (e.g., in some medical applications false positives are acceptable whereas false negatives could be fatal), it is also possible to use more detailed measures such as sensitivity, specificity, or precision to assess the discrimination capability of a component.
We also may consider the class priors.

The run-time complexity required to evaluate the discrimination of one component is $\mathcal{O}(N \cdot I \cdot \lvert\mathcal{C}\rvert \cdot D_{\mathrm{cont}}^3)$ since for each training sample we have to evaluate the density of each component for each class.

\subsubsection{Representativity} \label{sec:Representativity}

The performance of a generative classifier also depends on how well it models the data.
This kind of fitness is determined by the continuous dimensions only where we explicitly assume that the data distribution can be modeled by a mixture of Gaussians.
Since for categorical data it is always possible to find a distribution that perfectly models the data (cf. determination of a histogram), the \measure{representativity} measure only considers the continuous dimensions $\mathbf{x}^\textrm{cont}$.
As the true underlying distribution $q(\mathbf{x}^\textrm{cont})$ is unknown for real-world data sets, it must be approximated with a non-parametric density estimation technique, e.g., a standard \emph{Parzen window} density estimator:
\begin{equation}
  \begin{split}
    q(\mathbf{x}^\textrm{cont}) = \frac{1}{N} \! \sum_{\mathbf{x}_n \in \mathbf{X}} \frac{1}{\left(2 \pi h^2\right)^{\frac{D_\textrm{cont}}{2}}} \cdot\exp\left( - \frac{\|\mathbf{x}^\textrm{cont} - \mathbf{x}^\textrm{cont}_n\|^2}{2 \cdot h^2} \right).
  \end{split}
\end{equation}
Here, $h$ is a user-defined parameter. Suitable values of $h$ depend on the data set $\mathbf{X}$ \cite{Bis06}, but there are a number of heuristics to estimate $h$.
In \cite{CMB94}, for instance, $h$ is set to the average distance of the ten nearest neighbors for each sample, averaged over the whole data set.
This non-parametric approach makes no assumptions about the functional form of the underlying distribution.

The calculation of the representativity of the classifier is based on a divergence measure. Here, we rely on a variant of the Kullback-Leibler divergence $\mathrm{KL}(p_1(\mathbf{x}) || p_2(\mathbf{x}))$ which for two distributions $p_1(\mathbf{x})$ and $p_2(\mathbf{x})$ is defined as
\begin{equation}
  \mathrm{KL}(p_1(\mathbf{x}) || p_2(\mathbf{x})) = -\!\int p_1(\mathbf{x}) \ln\frac{p_2(\mathbf{x})}{p_1(\mathbf{x})} \dif\mathbf{x}.
\end{equation}
Other divergence measures, for example, Jensen–Shannon divergence, could also be used.
Since the measure is not symmetric, i.e.,~$\mathrm{KL}(p_1(\mathbf{x}) || p_2(\mathbf{x})) \neq \mathrm{KL}(p_2(\mathbf{x}) || p_1(\mathbf{x}))$, we use a variant which we denote $\mathrm{KL}_2(p_1(\mathbf{x}), p_2(\mathbf{x}))$.
It is given by:
\begin{align}
  \mathrm{KL}_2(p_1(\mathbf{x}),p_2(\mathbf{x})) =
  \frac{1}{2}\big(\mathrm{KL}(p_1(\mathbf{x}) || p_2(\mathbf{x})) + \mathrm{KL}(p_2(\mathbf{x}) || p_1(\mathbf{x}))\big). \label{eqn:kl2}
\end{align}
This measure always takes values greater than or equal to $0$ and only vanishes if $p_1(\mathbf{x})$ and $p_2(\mathbf{x})$ are identical.
It would also be possible to use the Hellinger distance $\Hel(p(\mathbf{x}^\textrm{cont}),q(\mathbf{x}^\textrm{cont}))$, cf.~Eq.~\eqref{eq:hellinger}, to measure the distance between the true data distribution $q(\mathbf{x}^\textrm{cont})$ and the model $p(\mathbf{x}^\textrm{cont})$, cf.\ Eq.~\eqref{eqn:px} and \eqref{eqn:gauss}.
However, since this measure is restricted to the unit interval, errors of the approximation in Eq.~\eqref{eqn:kl2approx} may result in values close to 1 and, thus, the difference between the models with and without a certain component is typically close to zero.
This effect is alleviated by using  $\mathrm{KL_2}$ which has no upper bound and, thus, the influence of a component on the representativity of the model can be quantified with reasonable precision even in the presence of approximation errors.

For the given distribution types Eq.~\eqref{eqn:kl2} cannot be solved analytically.
However, given a dataset $\mathbf{X}$ whose elements are distributed according to $p_1(\mathbf{x})$, the  $\mathrm{KL_2}$ divergence can be approximated as follows (cf.\ the concept of \emph{importance sampling})

\begin{align}
  \widehat{\mathrm{KL_2}}(p_1(\mathbf{x}),p_2(\mathbf{x})) \approx
  \frac{1}{2 N} \left(
  \sum_{\mathbf{x}_n \in \mathbf{X}} \ln \frac{p_1(\mathbf{x}_n)}{p_2(\mathbf{x}_n)} + \sum_{\mathbf{x}_n \in \mathbf{X}} \frac{p_2(\mathbf{x}_n)}{p_1(\mathbf{x}_n)} \ln \frac{p_2(\mathbf{x}_n)}{p_1(\mathbf{x}_n)}
  \right).
  \label{eqn:kl2approx}
\end{align}

\measure{Representativity} evaluates the influence of a component on the ``goodness of fit'' of the model regarding the data distribution.
To calculate the representativity of component $i$ we again create a second CMM without $i$ as described for the discrimination measure.
Then, we compare the symmetric $\widehat{\mathrm{KL}_2}$ distance of the CMM with ($p_\text{with}(\mathbf{x}^\textrm{cont})$) and without ($p_\text{without}(\mathbf{x}^\textrm{cont})$) component $i$:
\begin{align}
  \mathrm{\measureabrv{repr}}(i) := \widehat{\mathrm{KL}_2}(p_\text{without}(\mathbf{x}^\textrm{cont}), q(\mathbf{x}^\textrm{cont})) - \widehat{\mathrm{KL}_2}(p_\text{with}(\mathbf{x}^\textrm{cont}), q(\mathbf{x}^\textrm{cont})).
\end{align}

To assess the representativity of a whole classifier we could again use a weighted average of the representatvity values of the contained components.
Alternatively we could directly use
\begin{align}
  \widehat{\mathrm{KL_2}}(p_{\textrm{with}}(\mathbf{x}^{\textrm{cont}}), q(\mathbf{x}^{\textrm{cont}}))
\end{align}
as an assessment of the representativity of the whole classifier.

The run-time complexity required to evaluate the representativity of one component is $\mathcal{O}(N \cdot I \cdot D_{\mathrm{cont}}^3)$ since we have to evaluate the density of each sample which involves a matrix multiplication.

\subsubsection{Uncertainty} \label{sec:Uncertainty}

The parameters of the CMM are estimated in a Bayesian fashion (cf.\ Section \ref{subsec:cmm}), i.e., they are regarded as random variables whose distributions must be determined from sample data.
For the parameters of the categorical dimensions $\boldsymbol{\delta}_{d_i}$ the corresponding distributions are \emph{Dirichlet} distributions.
The centers $\boldsymbol{\mu}_i$ and covariance matrices $\boldsymbol{\Sigma}_i$ of the continuous dimensions are modeled with \emph{Gaussian-Inverse-Wishart} distributions \cite{Bis06}.

To quantify the \measure{uncertainty} of these parameter estimates we use the \emph{entropy} ${\cal H}$ \cite{Bis06}.
The entropy of a continuous random variable $\mathbf{x}$ with density $p(\mathbf{x})$ is
\begin{equation}
{\cal H}[\mathbf{x}] = -\!\int p(\mathbf{x}) \ln  p(\mathbf{x}) \dif\mathbf{x}.
\end{equation}
The more ``concentrated'' the probability mass of the distribution $p(\mathbf{x})$ is, i.e., the more certain the parameter estimate is, the lower is the entropy ${\cal H}[\mathbf{x}]$.
Its value is unbounded and can even be negative for continuous variables.

It is also possible to measure the uncertainty for every model parameter individually.
In this work, however, we want to calculate an aggregated value quantifying the uncertainty of the estimation of component $i$.
Thus, we sum up the entropies of the corresponding parameter distributions:
\begin{equation}
  \mathrm{\measureabrv{unct}}(i) := {\cal H}[\boldsymbol{\mu}_i, \boldsymbol{\Sigma}_i] + \sum_{\mathclap{d=D_\textrm{cont}+1}}^{D} {\cal H}[\boldsymbol{\delta}_{d_i}] \label{eq:unct}
\end{equation}
This summation of entropies naturally arises from the joint parameter distribution due to the assumption that the continuous dimensions are independent from the categorical ones and the categorical dimensions are mutually independent.
Since the absolute value of the entropy values of the categorical dimensions depends on the number of categories and the sum of entropies of all categorical dimensions depends on the number of dimensions in applications where those numbers are very different it may be desirable to weight the summands in Eq.~\eqref{eq:unct} with factors depending on the number of categories per dimension and/or the number of categorical dimensions.

Note that we do not consider the mixing coefficients here.
There is only one distribution for the whole classifier and, thus, the same entropy value would be added to every component.
The entropies of the mentioned distributions are
\begin{align}
  \allowdisplaybreaks
  \begin{split}
    \mathcal{H}[\boldsymbol{\mu}_i,\boldsymbol{\Sigma}_i] = {}& -\frac{D_\text{cont}}{2}\log\beta_i + \frac{D_\text{cont}(D_\text{cont}+1)}{4}\log\frac{\pi}{4} - \frac{D_\text{cont}}{2}\log(\mdet{\bs{W}_i}) + \frac{D_\text{cont}(\nu_i+1)}{2} \\
    &- \frac{\nu_i + D_\text{cont} + 2}{2}\sum_{d=1}^{D_\text{cont}}\psi\left(\frac{\nu_i +1 -d}{2}\right) + \sum_{d=1}^{D_\text{cont}}\log \Gamma\left(\frac{\nu_i +1 -d}{2}\right)
  \end{split}\\
  \begin{split}
    \mathcal{H}[\boldsymbol{\delta}_{d_i}] = {}& -\sum_{k=1}^{K_d} (\epsilon_{{d_k}_i} - 1) \big(\psi(\epsilon_{{d_k}_i}) - \psi(\widehat{\epsilon_{d_i}})\big) - \ln C(\boldsymbol{\epsilon}_{d_i}),
  \end{split}
\end{align}
cf.\ \cite{Bis06,Ji06}, for instance.
Here $\boldsymbol{\epsilon}_{d_i} = (\epsilon_{{d_1}_i},\dots,\epsilon_{{d_{K_d}}_i})$ is the parameter vector of the second-order Dirichlet distribution for categorical dimension $d_i$ and
\begin{align}
  \widehat{\epsilon_{d_i}} &= \sum_{k=1}^{K_d} \epsilon_{{d_k}_i}, \\
  C(\boldsymbol{\epsilon}_{d_i}) &= \frac{\Gamma(\widehat{\epsilon_{d_i}})}{\Gamma(\epsilon_{{d_1}_i}) \dots \Gamma(\epsilon_{{d_{K_d}}_i})}.
\end{align}
$\Gamma(\cdot)$ is the gamma function defined by
\begin{align}
  \Gamma(x) = \int_0^\infty t^{x-1} \exp(-t) \dif t,
\end{align}
and $\psi(\cdot)$ is the digamma function defined by (cf. \cite{Bis06})
\begin{align}
  \psi(x) = \frac{\dif}{\dif x} \ln \Gamma(x).
\end{align}

In contrast to our previous measures this measure uses the second-order distributions which arise during VI training (cf. Section~\ref{sec:TrainingOfCMM}) as distributions over the actual parameters of our CMM.
More details on training models using second-order distributions can be found in \cite{Bis06, FS09, FKS11}.

To evaluate the uncertainty of a whole classifier we could again compute a weighted average of the uncertainty values of the contained components using the mixing coefficients.
Additionally, we might take the uncertainty of the second-order distribution which models the mixing coefficients into account. This uncertainty is modeled by the variances of these distributions.

The run-time complexity required to evaluate the uncertainty of one component is
\begin{align}
  \mathcal{O}\left(D_{\mathrm{cont}}^3 + D_{\mathrm{cat}} \cdot \max_{d \in \{1, \dots, D_{\mathrm{cat}}\}}\{K_d\}\right),
\end{align}
since we have to compute the determinant of a matrix for the continuous dimensions and sum over the categories of all categorical dimensions.

\subsubsection{Distinguishability} \label{sec:Distinguishability}

For the rule set represented by the classifier to be easily \textit{understandable} for a human expert each two rules should be easily distinguishable.
Since categorical dimensions are easily distinguishable by humans we only consider the continuous dimensions here.
To measure the distinguishability of two components $i$ and $i'$ for dimension $d \in \{1, \dots, D_{\text{cont}}\}$ we first project those components onto $d$.
This yields two univariate Gaussians $\varphi_{d_i}$ and $\varphi_{d_{i'}}$.
In order to restrict our \measure{distinguishability} measure to the unit interval we omit the normalizing coefficients of the projected Gaussians.
This also makes sure that strongly overlapping Gaussians which are not easily distinguishable for a human are assigned a low distinguishability value.
To assess the distinguishability of the two components $i$ and $i'$ with regard to dimension $d$ we now use the intersection point of the two projected Gaussians that lies between their centers which results in
\begin{align}
  \mathrm{\measureabrv{dsng}}_d(i, i') := 1 - \exp\left(-\frac{(\mu_{d_i} - \mu_{d_{i'}})^2}{2 (\sigma_{d_i} + \sigma_{d_{i'}})^2}\right)
\end{align}
with $\mathrm{\measureabrv{dsng}}(i,i') \in (0,1]$.
We now aggregate those values over all components $i'$ different from $i$ and over all continuous dimensions $d$ to get the overall distinguishability for component $i$ as
\begin{align}
  \mathrm{\measureabrv{dsng}}(i) = \min_{d \in \{1, \dots, D_{\text{cont}}\}}\left\{\min_{\substack{i \in \mathcal{I}\\i\neq i'}}\{\mathrm{\measureabrv{dsng}}_d(i, i')\}\right\}.
\end{align}
Values higher than $0.1$, for example, could be regarded as desirable, depending on the application.
Fig.~\ref{fig:dsng} shows an example of the assessment of the distinguishability of two exemplary Gaussians.

\begin{figure}[htb]
  \centering
  \includegraphics{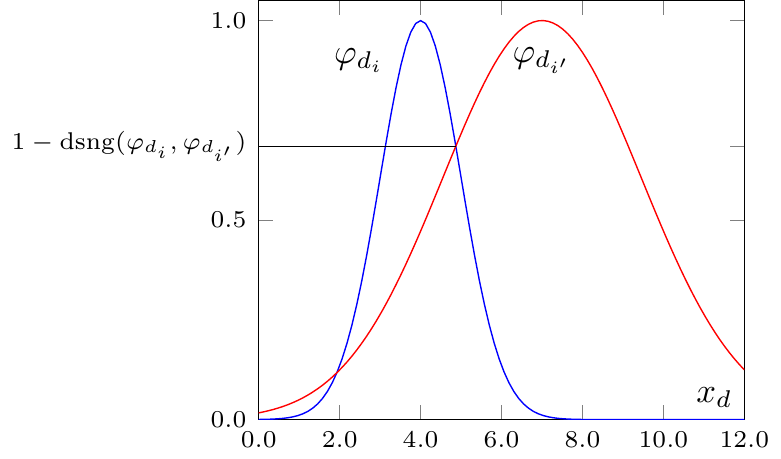}
  \caption{Example of an assessment of the distinguishability of two Gaussians.}
  \label{fig:dsng}
\end{figure}

The run-time complexity required to evaluate the \measure{distinguishability} of one component is $\mathcal{O}(D_{\mathrm{cont}} \cdot I)$ since for each continuous dimension we have to iterate over all components of the classifier to compute the intersection points of the projected Gaussians.

The distinguishability of a set of rules (i.e. the whole classifier) is defined to be the distinguishability of the rule with the lowest distinguishability.
When evaluating the complete rule set two additional factors may be considered regarding the interpretability of the rule set by a human expert.

First, the number $I$ of rules in the rule set should be low because we argue that a classifier with few rules is easier to understand than a classifier with many rules.

Second, the number of different terms $\tau_d$ for each dimension $d$ should be low.
For a categorical dimension $\tau_d$ is given by the number of different categories with non-zero probability forming the disjunctions.
To simplify categorical dimensions only categories with a probability above a certain threshold $\omega > 0$ may be considered (cf. Section~\ref{sec:RuleExtraction}).
Let $\mathcal{K}_{d_i}$ be the set of categories of dimension $d$ of component $i$.
Then we have
\begin{align}
  \tau_d = \left\lvert \bigcup_{i=1}^{I} \mathcal{K}_{d_i} \right\rvert.
\end{align}
For a continuous dimension $d$ the number of different univariate Gaussians $\varphi_{d_i}$ is counted.
To decide whether two Gaussians should be regarded as being different or not, we use the Hellinger distance (cf. Eq.~\eqref{eq:hellinger}) of the two Gaussians.
The distance should be clearly below a fixed threshold, such as $0.01$, for example, to regard two Gaussians as being identical:
\begin{align}
  \tau_d = \sum_{i=1}^{I} h_{d_i}
\end{align}
where
\begin{align}
  h_{d_i} =
  \begin{cases}
    0, & \begin{tabular}{@{}l@{}}if there is a $\varphi_{d_{i'}}$ with $i < i'$ and $\Hel(\varphi_{d_i}, \varphi_{d_{i'}}) \leq 0.01$ \end{tabular} \\
    1, & \text{otherwise}
  \end{cases}.
\end{align}
The threshold can be varied in applications depending on the degree of distinguishability that is desired.
To assess a complete rule set we could average the number of different terms for all categorical and continuous dimensions.
For the classifier shown in Fig.~\ref{fig:rules} we get $\tau_1 = 2$, $\tau_2 = 2$, and $\tau_3 = 3$ resulting in $\tau = 2.3$, for example.

\section{Case Studies} \label{sec:CaseStudies}

In this section we investigate the properties of the proposed measures in detail by (1) analyzing correlations between these measures and run-times, and by (2) conducting four case studies that demonstrate how the measures can be used in practical applications.
These case studies show how measures can be used in the learning phase of the classifier, to improve the classification performance in an active learning setting, while evaluating the trained classifier before using it on-line, and finally during the application phase of the classifier.
These case studies can be seen as illustrative examples; many other ways of using the measures are possible.

\subsection{Correlation Analysis and Run-Time} \label{sec:CorrelationAnalysis}

In the first set of experiments we analyze correlations between the seven measures to investigate their dependencies.
Additionally, we measure the run-time of all evaluations of our measures required to compute those correlations to get some empirical evidence to back up the theoretical run-times stated in Section~\ref{sec:Measures}.
For this experiment we again use the 21 benchmark data sets.

\begin{table}
  \centering\scriptsize
  \caption{Spearman’s correlation coefficients averaged for the seven interestingness measures over 21 benchmark data sets (with standard deviations). Entries which are statistically significant for a significance level of 0.05 are highlighted.\label{tab:Spearman}}
  \begin{tabular}{lrrrrrrr}
    \toprule
    & \multicolumn{1}{c}{\measureabrv{info}} & \multicolumn{1}{c}{\measureabrv{impo}} & \multicolumn{1}{c}{\measureabrv{uniq}} & \multicolumn{1}{c}{\measureabrv{disc}} & \multicolumn{1}{c}{\measureabrv{repr}} & \multicolumn{1}{c}{\measureabrv{unct}} & \multicolumn{1}{c}{\measureabrv{dsng}} \\
    \midrule
    \measureabrv{info} & 1.00${}\pm{}$0.00 & $-$0.23${}\pm{}$0.39 & \fontseries{b}\selectfont 0.56${}\pm{}$0.38 & 0.07${}\pm{}$0.28 & $-$0.14${}\pm{}$0.44 & 0.19${}\pm{}$0.44 & $-$0.01${}\pm{}$0.37 \\
    \measureabrv{impo} &  & 1.00${}\pm{}$0.00 & $-$0.19${}\pm{}$0.36 & \fontseries{b}\selectfont 0.55${}\pm{}$0.27 & \fontseries{b}\selectfont 0.56${}\pm{}$0.28 & \fontseries{b}\selectfont $-$0.51${}\pm{}$0.69 & 0.24${}\pm{}$0.34 \\
    \measureabrv{uniq} &  &  & 1.00${}\pm{}$0.00 & $-$0.01${}\pm{}$0.26 & $-$0.13${}\pm{}$0.47 & 0.24${}\pm{}$0.40 & 0.09${}\pm{}$0.34 \\
    \measureabrv{disc} &  &  &  & 1.00${}\pm{}$0.00 & 0.34${}\pm{}$0.28 & $-$0.18${}\pm{}$0.42 & 0.09${}\pm{}$0.33 \\
    \measureabrv{repr} &  &  &  &  & 1.00${}\pm{}$0.00 & $-$0.17${}\pm{}$0.57 & 0.18${}\pm{}$0.30 \\
    \measureabrv{unct} &  &  &  &  &  & 1.00${}\pm{}$0.00 & $-$0.07${}\pm{}$0.44 \\
    \measureabrv{dsng} &  &  &  &  &  &  & 1.00${}\pm{}$0.00 \\
    \bottomrule
  \end{tabular}
\end{table}

Table~\ref{tab:Spearman} shows Spearmans’s rank correlations computed for our seven interestingness measures and averaged over the 21 data sets.
Correlations that are statistically significant on a significance level of 0.05 are highlighted with bold typeface in Table~\ref{tab:Spearman}.
Those significant correlations are:
\begin{itemize}
\item \measure{Informativeness} is positively correlated to \measure{uniqueness}.
  This means that isolated components do not cover many samples.
  Furthermore, components that are located relatively close to other components of a CMM cover a majority of the samples.
  This may be due to the fact that many of the benchmark data sets consist of real-world data for which the normal distribution assumption may not be fully satisfied.
  This leads to clusters being modeled by multiple components rather than by one single component.
  It is also consistent with importance being negatively correlated to uncertainty.
\item \measure{Importance} is positively correlated to \measure{discrimination}, i.e., components that cover many samples also have a high impact on the decision boundary of the classifier.
\item \measure{Uncertainty} is negatively correlated to \measure{importance} which means that components covering only very few samples yield only very little additional information which expressed through the entropy yields the uncertainty.
\item \measure{Representativity} is positively correlated to \measure{importance}.
This is due to the fact that components that have a high mixing coefficient in comparison with the other components of the model also have a high influence on the overall density function which is evaluated by the \measure{representativity} measure.
\end{itemize}
For several combinations of interestingness measures the values of the correlation coefficients are quite.
Especially, the \measure{distinguishability} measure has no significant correlation (considering Spearman’s correlation coefficient) with any of the other measures.
This shows that the \measure{distinguishability} measure evaluates completely different aspects of the components of a CMM classifier than the other measures.
For example, a component can have a high \measure{distinguishability} while still having either a small or big influence on the decision boundary as measured by \measure{discrimination}.
Apart from the \measure{distinguishability}, the pair of measures with the lowest correlation is \measure{discrimination} and \measure{uniqueness}.
If a component has a very high \measure{uniqueness} it mostly “covers” samples of one class and if its \measure{uniqueness} is very low it “covers” samples of several classes.
In both cases the \measure{discrimination} of the component, i.e.\ its influence on the decision boundary, can either be very high or low.
This just depends on the position of the component with regard to its neighboring components and the decision boundary and notably not on the \measure{uniqueness} of the component.
Altogether, the fact that there are only a few significant correlations between our objective interestingness measures indicates that our measures cover many different aspects of knowledge understanding.

\begin{table}[htb]
  \centering\scriptsize
  \caption{Comparison of run-times in milliseconds required to compute the results in Table~\ref{tab:Spearman}. The last columns are the number of samples (N) and dimensions (D) of the data set and components (J) in the final model.}
  \label{tab:RunTime}
  \begin{tabular}{lrrrrrrrrrr}
    \toprule
               & \measureabrv{info}  & \measureabrv{impo}      & \measureabrv{uniq}   & \measureabrv{disc}    & \measureabrv{repr}     & \measureabrv{unct} & \measureabrv{dsng}  & N &  D & J \\
    \midrule
    australian & 198   & 95        & 2550   & 3013    & 163066   & 26   & 437  &  690  & 14 &  19 \\
    clouds     & 29    & 33        & 88     & 170     & 59940    & 14   & 30   &  5000 &  2 &  4  \\
    credit\_a  & 228   & 107       & 6703   & 6185    & 43692    & 65   & 99   &  690  & 15 &  26 \\
    credit\_g  & 3200  & 720       & 203792 & 216984  & 388329   & 407  & 1104 &  1000 & 20 & 107 \\
    ecoli      & 94    & 19        & 280    & 1020    & 7039     & 8    & 14   &  336  &  6 &  22 \\
    glass      & 207   & 72        & 394    & 970     & 5892     & 6    & 87   &  214  &  9 &  21 \\
    heart      & 5     & 3         & 94     & 70      & 1398     & 15   & 2    &  270  & 13 &  5  \\
    iris       & 7     & 1         & 32     & 86      & 719      & 1    & 26   &  150  &  4 &  7  \\
    pendigits  & 8760  & 8229      & 104277 & 1541128 & 18217114 & 3058 & 6076 & 10992 & 16 & 104 \\
    phoneme    & 844   & 334       & 17773  & 14708   & 1593644  & 256  & 415  &  5404 &  5 &  47 \\
    pima       & 851   & 190       & 4526   & 3993    & 109913   & 68   & 92   &  768  &  8 &  41 \\
    quality    & 10254 & 3570      & 294072 & 652816  & 18145798 & 1667 & 4342 &  4898 & 11 & 182 \\
    ripley     & 6     & 4         & 22     & 34      & 4214     & 4    & 86   &  1250 &  2 &  4  \\
    satimage   & 220   & 147       & 1075   & 4341    & 535105   & 41   & 98   &  6435 &  5 &  26 \\
    seeds      & 1     & ${}<{}$ 1 & 4      & 4       & 583      & 6    & 1    &  210  &  7 &  3  \\
    segment    & 2125  & 953       & 34704  & 136056  & 3133814  & 416  & 4643 &  2310 & 18 &  56 \\
    two\_moons & 482   & 291       & 1896   & 2919    & 1434223  & 111  & 159  & 14977 &  2 &  11 \\
    vehicle    & 5055  & 1688      & 32658  & 66724   & 615648   & 337  & 1735 &  846  & 18 &  60 \\
    vowel      & 11493 & 988       & 73744  & 154656  & 862679   & 426  & 3126 &  990  & 10 & 187 \\
    wine       & 499   & 192       & 869    & 781     & 8258     & 10   & 187  &  178  & 13 &  30 \\
    yeast      & 3638  & 492       & 35094  & 150671  & 606798   & 202  & 489  &  1484 &  6 &  57 \\
    \bottomrule
  \end{tabular}
\end{table}

We also measured the empirical run-times it took to compute the results given in Table~\ref{tab:Spearman}.
The results are presented in Table~\ref{tab:RunTime} and were obtained using a dedicated Linux machine with an Intel Core i7 2600 CPU running at 3.4 GHz.
The run-time measurements are predominantly a rough guideline to estimate how long it takes to evaluate individual measures in actual applications. In our experimental implementation we optimize many computations and rely on as many caches as possible.
For example, inverse covariance matrices required to evaluate the densities are already pre-computed by our training algorithm.
Thus, in contrast to what is suggested by the theoretical run-time complexities of our interestingness measures as stated in Section~\ref{sec:Measures}, the empirical run-times are not dominated by matrix operations but rather by the number of samples in each data set.
For small data sets such as seeds, ripley, iris, or heart, the evaluations of all measures require less than a second.
In comparison, the pendigits and quality data sets require the highest run-times up to a few hours.
With regard to the measures, representativity generally takes the longest to evaluate, followed by discrimination and uniqueness.
The evaluation of the other measures is very fast.

\subsection{Knowledge Acquisition Phase: Controlling the Training of a Classifier} \label{sec:CaseStudy1}

In this set of experiments we use some of our measures, \measure{importance}, \measure{representativity}, and \measure{uncertainty}, to control the VI based training process of a classifier.
\begin{figure}%
  \centering%
  \subcaptionbox{Average classification error on test data with standard deviation for pruning method Resp.}{%
    \includegraphics{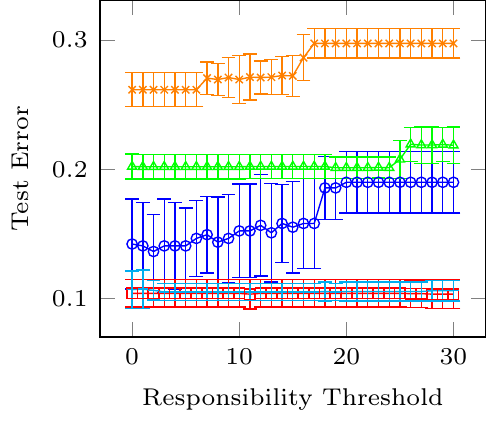}%
  }%
  \hfill%
  \subcaptionbox{Average number of finally resulting components with standard deviation for pruning method Resp.}{%
    \includegraphics{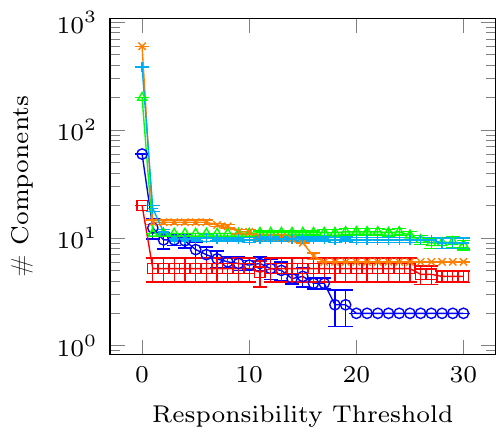}%
  }%
  \hfill%
  \subcaptionbox{Average number of VI steps with standard deviation for pruning method Resp.}{%
    \includegraphics{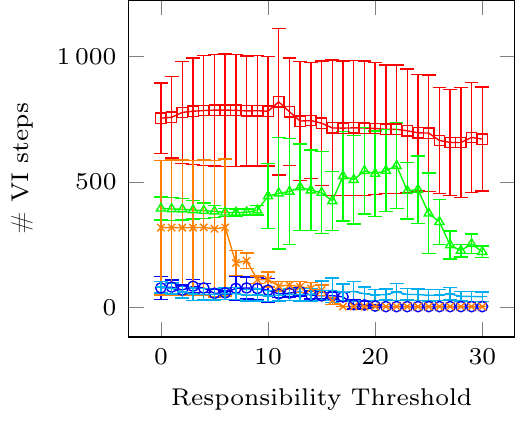}%
  }
  \par\bigskip%
  \subcaptionbox{Average classification error on test data with standard deviation for pruning method Impo.}{%
    \includegraphics{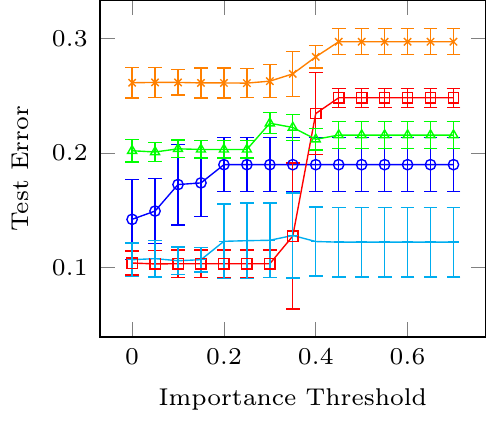}%
  }%
  \hfill%
  \subcaptionbox{Average number of finally resulting components with standard deviation for pruning method Impo.}{%
    \includegraphics{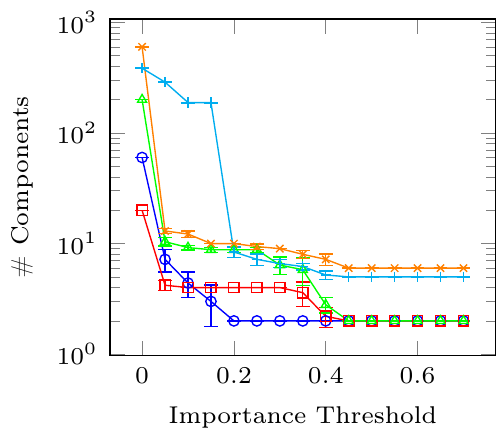}
  }%
  \hfill%
  \subcaptionbox{Average number of VI steps with standard deviation for pruning method Impo.}{%
    \includegraphics{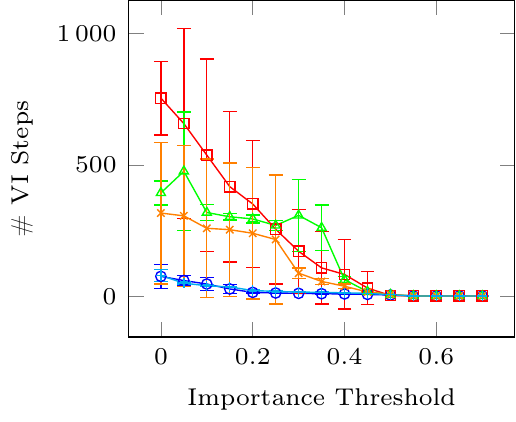}%
  }%
  \par\bigskip%
  \subcaptionbox{Average classification error on test data with standard deviation for pruning method Unct.}{%
    \includegraphics{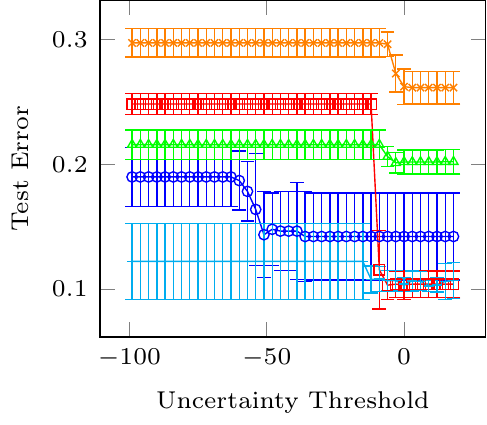}%
  }%
  \hfill%
  \subcaptionbox{Average number of finally resulting components with standard deviation for pruning method Unct.}{%
    \includegraphics{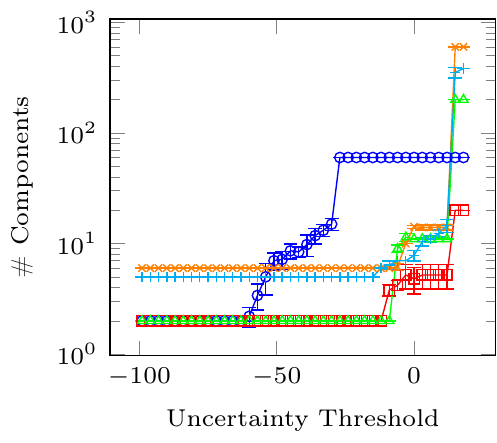}%
  }%
  \hfill%
  \subcaptionbox{Average number of VI steps with standard deviation for pruning method Unct.}{%
    \includegraphics{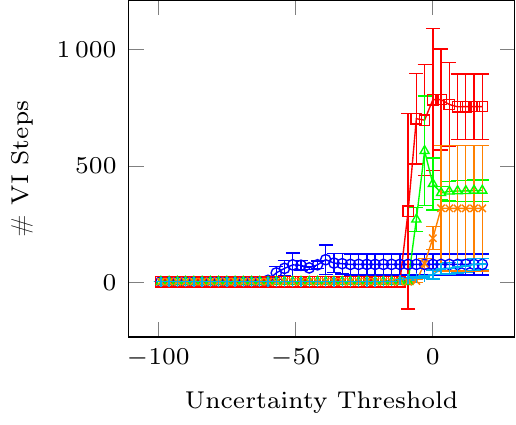}%
  }%
  \par\bigskip%
  \subcaptionbox{Legend specifying the $5$ used benchmark data sets.}{%
    \includegraphics{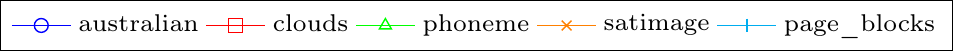}%
  }%
  \caption{Comparison of different pruning methods applied to  five benchmark data sets using a $5$-fold cross-validation. Data was first reduced to two dimensions using a PCA and then normalized using a $z$-transform. Note that in images (a) to (f), components are pruned if the measure drops \textit{below} the given threshold (horizontal-axis), whereas in images (g) to (i) components are pruned if the given threshold is \textit{exceeded}.}
  \label{fig:pruning}
\end{figure}
The training consists of three steps, an E step, an M step and an additional pruning step, which are consecutively executed until a convergence criterion is met.
In the E step, samples are gradually assigned to components of the classifier and in the M step, the parameters of the classifier are updated according to the samples (gradually) assigned to them.
In the pruning step, unnecessary components are removed from the classifier.
Here, we focus on the pruning step and consider three different pruning methods:
\begin{enumerate}
\item \textbf{Resp:} This method is similar to the ``traditional'' pruning method used in VI training: A component is removed from the classifier if the sum of its un-normalized responsibilities (cf.~Eq.~\eqref{eq:Responsibilities}) is \textit{below} a certain threshold.
  Often, this threshold is chosen as $1$ which means that a component is removed if it is effectively ``responsible'' for less than one sample of the training data.
  A threshold of $2.5$ means the component is ``responsible'' for less than $2.5$ samples and so on.
  A threshold of $0$ is also possible in which case no pruning is done.
\item \textbf{Impo:} Based on the \measure{importance} measure (cf.~Section~\ref{sec:Importance}) this pruning method uses the mixing coefficient of a component to decide whether or not to prune it.
  However, in contrast to Resp not only one component is considered but the mixing coefficient is set in relation to the average mixing coefficient of the classifier (or, in other words, the number of components the classifier currently consists of).
  Thus, a component is removed if its importance is \textit{below} a certain threshold.
  Since importance values lie in the interval $[0,1]$, thresholds in that interval may be considered.
  Values close to $0$ result in slow pruning and values close to $1$ lead to fast pruning.
  A threshold of $0.5$ means that all components whose mixing coefficient is below average get pruned.
\item \textbf{Unct:} In contrast to Resp and Impo this method does not rely on mixing coefficients or responsibilities but takes a completely different approach.
  It considers the \measure{uncertainty} of a component (cf.~Section~\ref{sec:Uncertainty}) and removes all components whose uncertainty value lies \textit{above} a certain threshold.
  The uncertainty of a component is unbounded and it is, thus, more difficult to determine a suitable threshold.
  However, the following examples provide good starting points for choosing a threshold value.
\end{enumerate}

Fig.~\ref{fig:pruning} visualizes classification errors on test data, numbers of resulting classifier components, and numbers of required VI training steps for the three different pruning methods.
We used $4$ of the benchmark data sets from Section~\ref{sec:CorrelationAnalysis} for this experiment: australian, clouds, phoneme, and satimage.
Additionally, we added the page\_blocks data set from the UCI Machine Learning Repository \cite{AN07} which consists of real-world data.
It can be seen that the Resp method yields the best results for low thresholds in the range $[1,3]$ but still requires a larger number of VI steps than the other methods.
In contrast to Resp, the Impo pruning method leads to a faster decrease in the number of required VI steps while resulting in a similar test error and number of components.
The threshold used in the \textbf{Unct} method is rather sensitive to the type of data set used.
The australian data set contains categorical dimensions in addition to continuous ones and this leads to rather different entropy values and, thus, different thresholds.
In conclusion, we can state that importance and uncertainty can both be used as alternatives to responsibilities for a pruning step of the VI training algorithm by reducing the number of required VI steps and thus making the training process faster.
However, further research into an automatic determination of the pruning threshold and the behavior of the pruning methods on different data sets is necessary.

\subsection{Knowledge Analysis Phase: Ranking Components of a Trained Classifier} \label{sec:CaseStudy2}

In this experiment we analyze the components of a trained classifier with our measures in order to help a potential user of the classifier in understanding the components or rules extracted from those components (cf.\ Section~\ref{sec:RuleExtraction}).

We trained a classifier on the phoneme data set from the UCL Machine Learning Group \cite{UCL14} using the restriction to diagonal covariance matrices to enable rule extraction.
Fig.~\ref{fig:RankingClassifier} shows the resulting classifier.
Components are colored according to their class densities and their opacity is dependent on their mixing coefficient.
The background of the plot is colored according to the posterior probabilities of the classifier.
The black, solid line is the decision boundary.

\begin{figure}[htb]
  \centering
  \includegraphics[width=.5\textwidth]{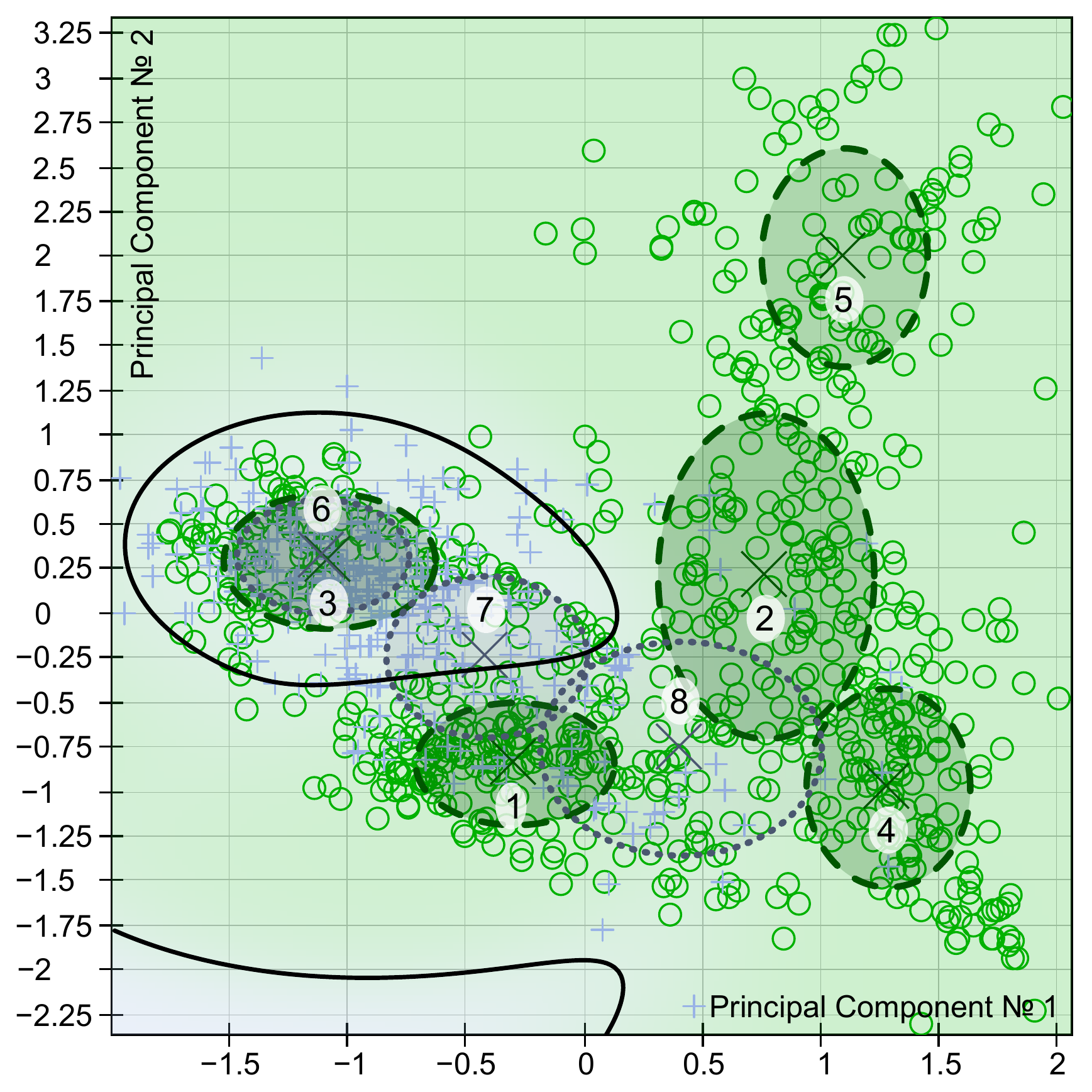}
  \caption{A fraction of the data of the phoneme data set together with a trained classifier. The data were reduced to two dimensions using a PCA. Components of the first class are visualized by dashed green ellipses and numbered below their center. Components of the second class are shown as dotted blue ellipses and are numbered above their center.}
  \label{fig:RankingClassifier}
\end{figure}

In Table~\ref{tab:ranking}, some of our measures are evaluated for the components depicted in Fig.~\ref{fig:RankingClassifier}.
Other measures were omitted for the sake of brevity.

\begin{table}[htb]
  \centering\scriptsize
  \caption{Ranking of components based on various measures. Component numbers are given in Fig.~\ref{fig:RankingClassifier}.}
  \label{tab:ranking}
  \begin{tabular}{lrcc}
    \toprule
    Measure & \multicolumn{1}{c}{Value} & Rank & Component \\
    \midrule
    \multirow{8}{*}{\measureabrv{dsng}} & 0.0022 & 1 & 3 \\
    & 0.0022 & 2 & 6 \\
    & 0.0037 & 3 & 2 \\
    & 0.0134 & 4 & 1 \\
    & 0.0134 & 5 & 8 \\
    & 0.0498 & 6 & 4 \\
    & 0.0583 & 7 & 7 \\
    & 0.2574 & 8 & 5 \\
    \midrule
    \multirow{8}{*}{\measureabrv{disc}} & $-$0.0028 & 1 & 3 \\
    & 0.0000 & 2 & 4 \\
    & 0.0000 & 3 & 5 \\
    & 0.0028 & 4 & 8 \\
    & 0.0157 & 5 & 7 \\
    & 0.0194 & 6 & 2 \\
    & 0.0722 & 7 & 6 \\
    & 0.1156 & 8 & 1 \\
    \bottomrule
  \end{tabular}
  \hspace{1cm}
  \begin{tabular}{lrcc}
    \toprule
    Measure & \multicolumn{1}{c}{Value} & Rank & Component \\
    \midrule
    \multirow{8}{*}{\measureabrv{info}} & 0.1342 & 1 & 3 \\
    & 0.1342 & 2 & 6 \\
    & 0.5128 & 3 & 1 \\
    & 0.5128 & 4 & 7 \\
    & 0.5221 & 5 & 2 \\
    & 0.5221 & 6 & 8 \\
    & 0.6267 & 7 & 4 \\
    & 0.7401 & 8 & 5 \\
    \midrule
    \multirow{8}{*}{\measureabrv{uniq}} & 0.2081 & 1 & 8 \\
    & 0.3900 & 2 & 3 \\
    & 0.4515 & 3 & 7 \\
    & 0.6293 & 4 & 6 \\
    & 0.7815 & 5 & 1 \\
    & 0.9074 & 6 & 2 \\
    & 0.9482 & 7 & 4 \\
    & 0.9991 & 8 & 5 \\
    \bottomrule
  \end{tabular}
\end{table}

From the \measure{dsng} measure we can deduce that components 3 and 6 are those which can least easily be distinguished.
This can be confirmed by looking at Fig.~\ref{fig:RankingClassifier} which shows that those components overlap to a great extent.
Components 5 and 7 can best be distinguished from others when projected to the coordinate axes which is obvious especially in case of component 5 because it is well separated from all other components.

The second measure evaluated in Table~\ref{tab:ranking} is \measure{disc}.
Since the discrimination value of component 3 is negative, we know that the classification performance could be increased if we removed this component from the classifier.
In this case, the classification error on test data decreases from $21.46\%$ to $21.18\%$.
Also, components 4 and 5 are not very important for the classification decision since they are ``dominated'' by component 2 as we can see from Fig.~\ref{fig:RankingClassifier}.
The components that are most important for the classification decision are components 1 and 6.
Those are the only green colored components in an area otherwise dominated by the blue components 3, 7, and 8.

The informativeness measure info in Table~\ref{tab:ranking} yields components 4 and especially 5 as those with the greatest distance to all other components.
In fact, they have nearly no overlap with other components in Fig.~\ref{fig:RankingClassifier}.
The least informative components are components 3 and 6 since they almost totally overlap.

The final measure we evaluated in Table~\ref{tab:ranking} is \measure{uniq}.
The components with the highest uniqueness are components 2, 4, and in particular 5 which means they mostly cover samples from one class.
Components 3 and 8 have the lowest uniqueness which means they cover samples from both classes.
Interestingly, component 6 which nearly completely overlaps with component 3 has a higher uniqueness value.

The example shows that the proposed measures can easily be used to analyze a trained classifier.
It should be emphasized that the evaluation of a classifier does also work as described in case of higher dimensional input spaces where looking at a two-dimensional visualization would not help very much.
Depending on the application, different goals could be achieved by ranking the components of a classifier using our approach.
If the aim was to reduce the number of components, for example, one might consider removing component 8 since it received the worst evaluation by four out of seven measures (not all shown here for the sake of brevity): it has the lowest uniqueness, representativity, and importance, and the highest uncertainty.

\subsection{Knowledge Analysis in Active Learning}

The pool-based active learning (PAL)~\cite{LG94} paradigm repetitively asks users (generally termed as \textit{oracles}) to provide label information for unlabeled data, e.g., in order to train a classifier based on those data. 
PAL is based on the assumption that unlabeled data can be acquired at no (or low) costs, whereas retrieving label information is very costly. 
Therefore, at the beginning of a PAL learning cycle a large set of unlabeled data and only a small set of labeled data are available.
Based on these data a classifier is trained (here a CMM), which is further on used to build a ranking of the unlabeled data based on an estimate how likely is it for a sample to increase the performance of the classifier if it was labeled. 
A \textit{selection strategy} uses these estimates as decision basis for selecting in each learning cycle the next sample or set of samples that is going to be labeled.
Fig.~\ref{fig:PalCycle} depicts the learning cycle of standard PAL with solid arrows.
Typically, a large pool of unlabeled samples ($U$) and a small set of labeled samples ($L$) are available at the beginning of PAL, which makes it possible for a classifier $\mathbf{G}$ to be trained.
Then, a set $S$ ($S \subset U$) is queried based on a selection strategy $\mathcal{Q}$ and presented to the oracle $\mathcal{O}$ for labeling.
The labeled samples ($S_{labeled}$) are added to $L$ and the classifier is updated. 
As long as a given stopping condition is not met, a new PAL cycle (i.e., learning cycle) is started.

The PAL process starts with a generative (probabilistic) model as described in Section~\ref{subsec:cmm} which can be trained in an unsupervised way.
During the PAL process, more and more labels become available that can be considered to train a CMM. 
In general, the underlying generative model remains unchanged during the PAL process. 
In~\cite{RCS15}, this process is extended with a \textit{transductive learner}, which aims at updating the underlying model using the \measure{uniqueness} measure, which determines how ambiguous the knowledge modeled by each component is.
If the uniqueness value determined for a component is smaller than a predefined threshold, the component is considered to be ``disputed'' between two or more classes.
In this case, the samples for which the component is responsible (cf.~Eq.~\ref{eq:Responsibilities}) are determined, a sample-based classifier called Resp-$k$NN~\cite{RC14} is trained and used to transductively label the underlying samples.
Then, a new CMM is trained and the resulting components \textit{fused} with ``non-disputed'' components, where necessary(for details of the fusion technique, see~\cite{FKS14}).
Consequently, the information provided by the uniqueness measure allows for iteratively improving a CMM during a PAL process.
For further information regarding PAL with a transductive learner see~\cite{RCS15}.
This extension of the PAL process is also shown in Fig.~\ref{fig:PalCycle}.

\begin{figure}[htb]
	\centering
	\includegraphics[width=.6\textwidth]{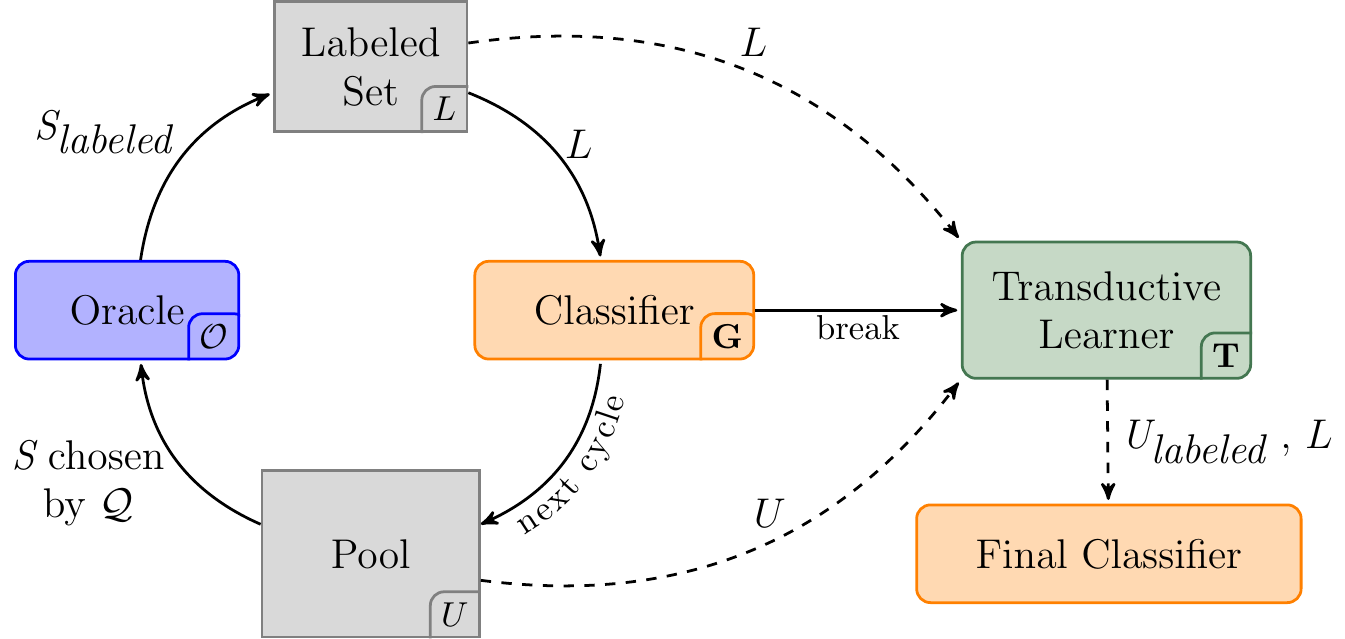}
	\caption{Learning cycle of PAL (solid arrows) with extension (dashed arrows).}
	
	\label{fig:PalCycle}
\end{figure}

\begin{figure}[htb]
	\centering
	\includegraphics[width=.95\textwidth]{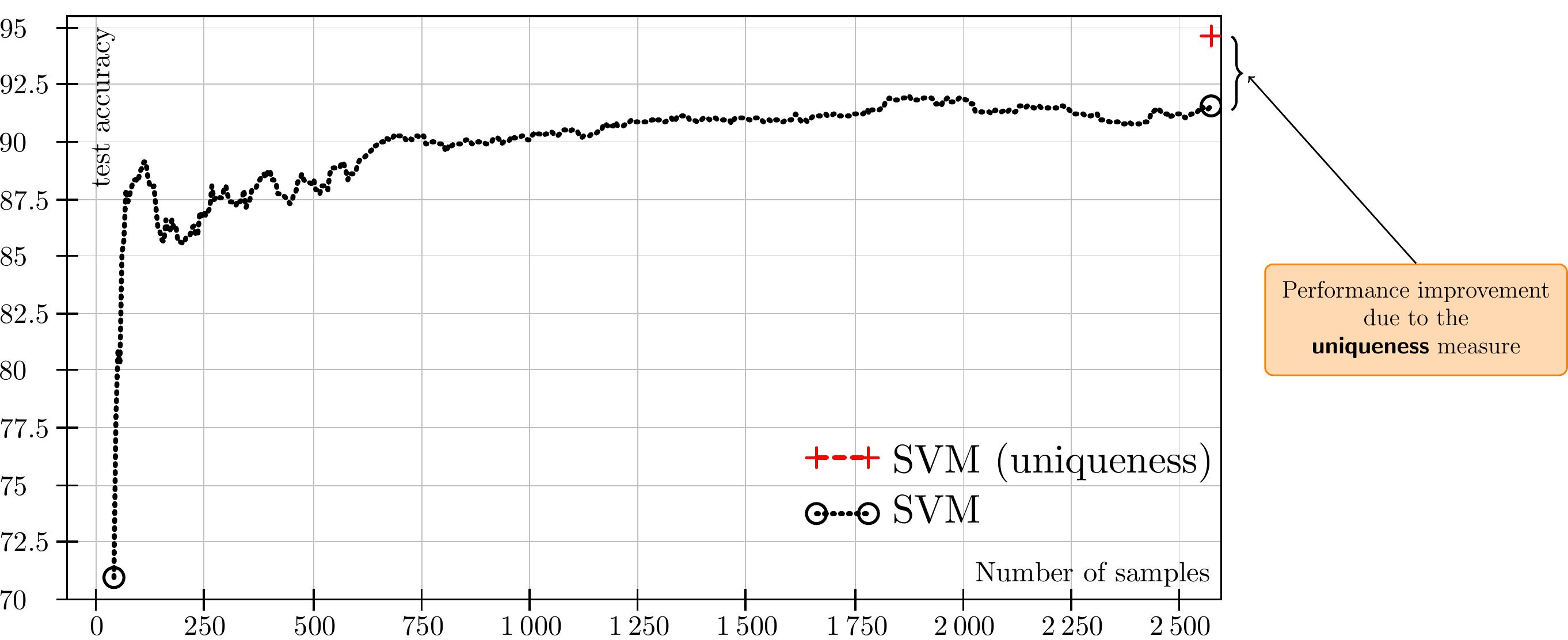}
	\caption{Learning curve (test accuracy versus the number of actively selected samples) of the active training of SVM with generative kernel (based on GMM) for the MNIST data set (magenta). The red $+$ indicates the test accuracy after refining the underlying GMM of the kernel by means of the \measureformat{uniqueness} measure.
			}
	\label{fig:RwmActive}
\end{figure}

For the case study presented in this section, we embed the probabilistic generative model in a kernel function of a support vector machine (SVM) as described in~\cite{RS15}.
In this case study, the \measureformat{uniqueness} measure is only applied once, after a final PAL cycle, to emphasize on the possible performance improvement.

We have conducted an experiment with the MNIST~\cite{MNIST15} handwritten digits data set, which consists of a training set of $60\,000$ and a test set of $10\,000$ gray-scaled images of handwritten digits ($10$~classes). 
We reduced the size of the input dimensions from $784$ to $34$ (continuous dimensions) by applying a principle component analysis, thus keeping $90\%$ of the total variance.
Furthermore, the data has been $z$-normalized. 
The PAL process started with an initially labeled set of $40$ samples and selected, based on the 4DS~\cite{RS13} selection strategy (corresponds to $\mathcal{Q}$ in Fig.~\ref{fig:PalCycle}) five samples in each learning cycle of the PAL process. 
We decided to actively select $2\,575$ samples (i.e., the PAL stopped when the size of the labeled set $L$ reached this number).
The actively trained classifier (here SVM) had been parametrized only using the training set.

After the PAL process stopped, an adequate number of labeled samples was available.
Thus, it was possible to improve the underlying density model by means of the \measureformat{uniqueness} measure as sketched above.
Then, the density model can better exploit the structure information contained in the labeled samples.
Fig.~\ref{fig:RwmActive} illustrates the learning curve of the SVM on the test set. 
Moreover, it emphasizes that by employing the \measureformat{uniqueness} measure the classification error is reduced by about a half.

This promising result indicates a possible improvement of the standard PAL process. 
In regular time intervals (i.e., cycles of the PAL process) the underlying density model should be evaluated by means of the \measureformat{uniqueness} measure and, if necessary, refined.

\subsection{Knowledge Application Phase: Detecting Novel Processes} \label{sec:CaseStudy3}

Our last set of experiments deals with the application phase of a classifier.
We consider the task of novelty detection, i.e., the task of detecting the need of new components that model newly occurring processes in the input space which were not known during training time (cf.\ \cite{MS13}).
Our example is based on intrusion detection data from the 1999 KDD cup \cite{KDD99}.
Specifically, we use data of the attack types ``neptune'', ``smurf'', ``ipsweep'', and ``satan''.
We re-sampled the data to be able to consider longer time spans.
Each of the four data sets we constructed starts with $54\,000$ samples of background traffic (large variety but without any attacks).
Then, an attack phase starts which lasts until time step $198\,000$.
During the attack phase samples that represent the respective attack are present in a ratio of 1:3.
Finally, we continue with more samples only drawn from background traffic to observe how the novelty detectors react to the end of an attack.
The classifiers were trained with a separate training data set consisting of $5\,000$ samples of background traffic only.
Our goal is to detect the need for a new component that models data originating from an attack to the network.
This shows that intrusion detection systems that detect new kinds of attacks at run-time can be build by relying on our proposed interestingness measures.

\begin{figure}[htb]
    \centering%
    \subcaptionbox{Representativity based novelty detection.\label{fig:NoveltyDetectionA}}{%
        \includegraphics{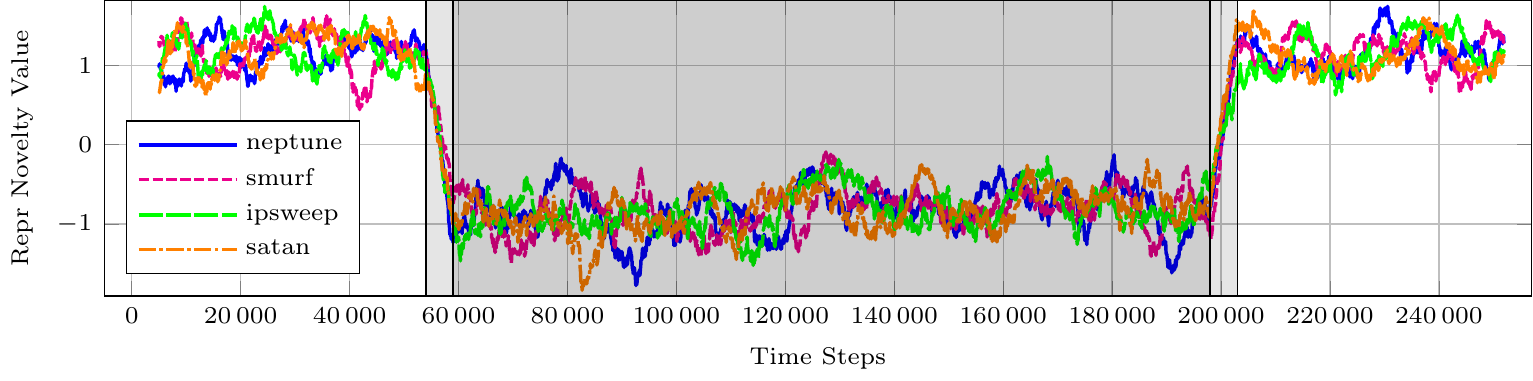}%
    }%
    \par\bigskip%
    \subcaptionbox{$\chi^2$ novelty detection.\label{fig:NoveltyDetectionB}}{%
        \includegraphics{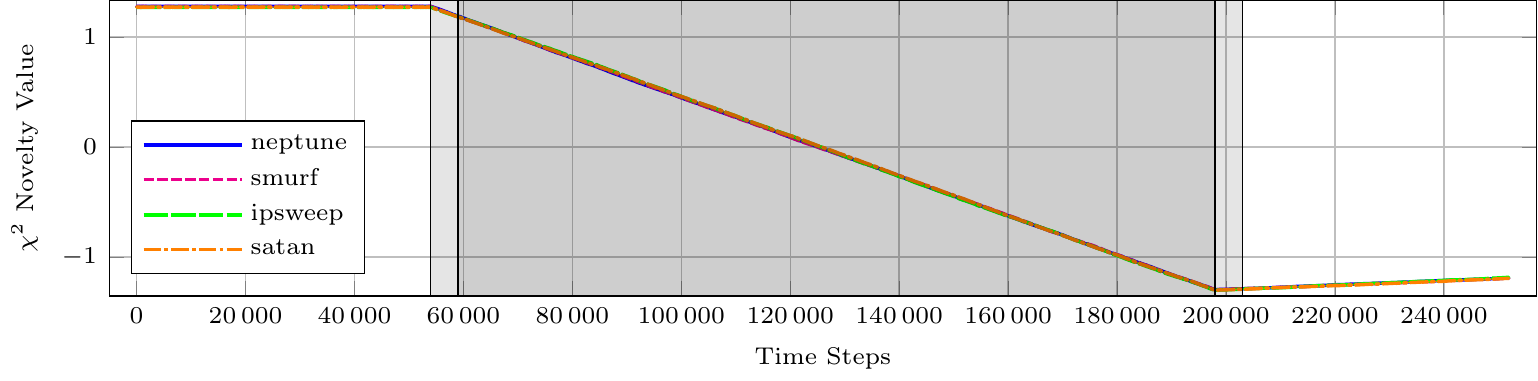}%
    }%
    \caption{Novelty detection with our representativity measure (top) in comparison to the $\chi^2$ novelty detector as described in \cite{Fis12} (bottom) on $4$ intrusion detection data sets. The grayed region marks the time an attack occurred. The light gray area indicates the size of the sliding window used by the representativity measure after the attack started and after it stopped again.}
    \label{fig:NoveltyDetection}
\end{figure}

In order to achieve that goal we evaluate the representativity of the overall classifier after each newly observed sample.
The Parzen window estimator of the \measure{representativity} measure that is used for the evaluation of repr is computed from a sliding window of the most recent $5\,000$ samples.
In Fig.~\ref{fig:NoveltyDetection}\,(\subref{fig:NoveltyDetectionA}), representativity values of the classifiers are shown.
For the sake of visualization, the representativity values were normalized to mean $0$ and variance $1$.
During the initial phase, which contains only background traffic, the representativity values of all classifiers are high.
When an attack starts at time step $54\,000$, the representativity values decrease until the sliding window is filled with data that contains 25\% of attack samples.
Now, the representativity values stay small until the attack to the network stops at time step $198\,000$.
Then, the representativities of the classifiers rise again to high values, similar to the ones in the initial phase.

For comparison, Fig.~\ref{fig:NoveltyDetection}\,(\subref{fig:NoveltyDetectionB}) contains results obtained by the novelty detection technique called $\chi^2$ novelty which is proposed in \cite{Fis12}.
This technique measures whether or not each newly observed sample lies within a certain region around any of the classifier components and accumulates reward or punishment values accordingly.
Fig.~\ref{fig:NoveltyDetection}\,(\subref{fig:NoveltyDetectionB}) depicts the accumulated novelty values.
For the sake of visualization those values were also normalized to mean $0$ and variance $1$.

In conclusion we can state that our \measure{representativity} measure can be used to detect the need for new components in a classifier.
The raw representativity is more sensitive to novel samples than the $\chi^2$ novelty measure which is more robust, but it is certainly worth to investigate techniques that combine the advantages of both.

\section{Conclusion and Outlook}\label{sec_conclusion}

In this article, we presented an approach to support or even automate the process of knowledge understanding, an important data mining step. However, the proposed measures should be seen as a first step into that direction, as not all aspects of real world applications could be considered yet.
With knowledge understanding we refer to the task of analyzing the knowledge extracted from data in order to gather meta-knowledge (knowledge about knowledge).
This can either be done offline (here referred to as ``understanding'') or online (here referred to as ``experience'') by applying the knowledge.
Depending on a concrete application, all or only a part of the measures may be useful. Also it may be preferable to assess continuous and categorical dimensions separately.

The measures proposed in this article are objective measures but they have some relation to subjective interestingness measures. For example, \measure{info} is clearly related to \textit{novelty}, a ``conventional'' \textit{interestingness measure}, since if a rule is quite informative it may also be the case that the user regards this rule as providing a highly novel kind of knowledge.
\measure{Uniq} can be seen as being related to \textit{understandability}, because a high \measure{uniq} value implies that a rule is only associated with a single class, thus making it better understandable.
The \measure{impo} and \measure{disc} measures are both reflecting the \textit{usefulness} of a component by expressing its relative weight in the classifier and the influence on the decision boundary. Another measure related to \textit{usefulness} is our \measure{unct}: A high value states that the parameters of the process underlying the resulting component cannot be precisely determined, which decreases its \textit{usefulness} of the component. The last measure \measure{repr}, which expresses how well a component models the data, has a positive effect on the \textit{understandability} of the overall model.

We defined various measures for CMM (classifier based on a mixture model), a generative, probabilistic classifier.
In practical applications, this classifier can be used in combination with discriminative classifiers such as support vector machines (SVM).
As we have shown in \cite{FKSO10}, the probabilistic classifiers with continuous input dimensions are \textit{functionally} equivalent to certain kinds of support vector machines (with Gaussian kernels), radial basis function neural networks, fuzzy classifiers (with Gaussian membership functions and sum-prod inference), nonlinear Fisher discrimination techniques, relevance vector machines, or direct kernel machines.
Thus, some of the measures defined here could be adapted to these kinds of classifiers, as well.
As the semantics underlying the knowledge in these classifiers is very different (cf., e.g., membership degrees in fuzzy systems to probabilities in our generative classifiers), this idea has to be investigated in detail.
Another option is to use the probabilistic model contained in our CMM in other classifiers, for instance, as we have already shown in \cite{RS15}, to create a data dependent kernel function for SVM.

In this article, several use cases were defined and investigated to demonstrate the value and applicability of the proposed measures.
We have shown that these measures may be used to control the training process of the classifier (for more details on this problem see also  \cite{Gru14}),
to analyze the knowledge contained in a trained classifier,
and to support tasks such as novelty detection in the application phase of the classifier.
Other applications are possible, e.g., pruning at the end of a training process, extraction and ranking of rules contained in a classifier (see, e.g., \cite{FKS11}), concept drift or obsoleteness detection.

In our future work, we will also focus on measures for a quantification of experience gained by applying the knowledge extracted from data (e.g., by assessing its usefulness as in \cite{FJKS12,TAAS2011}), extending the VI training techniques to apply them to large data sets, and on applications of the measures, e.g., in the field of collaborative intrusion detection in cyber-physical systems~\cite{secCPS2012}. Further, we will investigate how our measures can be adjusted to suite other types of distributions (i.e.,~distributions that are required for other attribute types) as well.

\section*{Acknowledgment}

This work was supported by the German Research Foundation (DFG)
under grant SI 674/9-1 (project CYPHOC).

\bibliographystyle{splncs03}
\bibliography{measures-arxiv}

\end{document}